\documentclass[lettersize,journal]{IEEEtran}
\usepackage{amsmath,amsfonts}
\usepackage{array}
\usepackage[caption=false,font=normalsize,labelfont=sf,textfont=sf]{subfig}
\usepackage{textcomp}
\usepackage{stfloats}
\usepackage{url}
\usepackage{verbatim}
\usepackage{graphicx}
\usepackage{soul,color}
\usepackage{bbm}
\usepackage{bm}
\usepackage{amsfonts,amssymb,amsmath}
\usepackage{booktabs}
\usepackage{multirow}

\usepackage[utf8]{inputenc}
\usepackage[linesnumbered,ruled,vlined]{algorithm2e}
\usepackage{enumitem}
\usepackage{xargs}                      % Use more than one optional parameter in a new commands
\usepackage[pdftex,dvipsnames]{xcolor}  % Coloured text etc.

\usepackage[dvipsnames,table,xcdraw]{xcolor}
\usepackage{tcolorbox}

\begin{document}

\title{A Physics-informed Demonstration-guided Learning Framework for Granular Material Manipulation}
\author{Minglun Wei$^{1}$, Xintong Yang$^{1}$, Yu-Kun Lai$^{2}$, Seyed Amir Tafrishi$^{1}$ and Ze Ji$^{1}$% <-this % stops a space
\thanks{Minglun Wei was supported by the UK Engineering and Physical Sciences Research Council (EPSRC) through a Doctoral Training Partnership (No. EP/W524682/1). This work was also partially supported by the UK EPSRC grant No. EP/X018962/1. Corresponding author: Ze Ji.}% <-this % stops a space
\thanks{$^{1}$Minglun Wei, Xintong Yang, Seyed Amir Tafrishi, and Ze Ji are with the School of Engineering, Cardiff University, Cardiff, CF24 3AA, United Kingdom.
        {\tt\small \{WeiM9, YangX66, TafrishiSA, JiZ1\}@cardiff.ac.uk}}%
\thanks{$^{2}$Yu-Kun Lai is with the School of Computer Science and Informatics, Cardiff University, Cardiff, CF24 4AG, United Kingdom.
        {\tt\small LaiY4@cardiff.ac.uk}}%
}

\markboth{Journal of \LaTeX\ Class Files,~Vol.~14, No.~8, August~2021}%
{Shell \MakeLowercase{\textit{et al.}}: A Sample Article Using IEEEtran.cls for IEEE Journals}


\maketitle

\begin{abstract}
Due to the complex physical properties of granular materials, research on robot learning for manipulating such materials predominantly either disregards the consideration of their physical characteristics or uses surrogate models to approximate their physical properties. Learning to manipulate granular materials based on physical information obtained through precise modelling remains an unsolved problem. In this paper, we propose to address this challenge by constructing a differentiable physics-based simulator for granular materials using the Taichi programming language and developing a learning framework accelerated by demonstrations generated through gradient-based optimisation on non-granular materials within our simulator, eliminating the costly data collection and model training of prior methods. Experimental results show that our method, with its flexible design, trains robust policies that are capable of executing the task of transporting granular materials in both simulated and real-world environments, beyond the capabilities of standard reinforcement learning, imitation learning, and prior task-specific granular manipulation methods.
\end{abstract}


\section{INTRODUCTION}\label{sec:intro}
\IEEEPARstart{P}{ouring} seasonings into a dish and adding sugar to coffee are routine actions in kitchen scenarios. For humans, manipulating such granular materials is effortless, owing to an inherent understanding of their physical properties. This knowledge enables humans to prevent spillage and accurately control the angle of the tools such as spoons to scoop or pour these substances. On the other hand, today's robots still struggle to understand the underlying physics and accomplish such delicate manipulation tasks. Indeed, whether in household kitchens, garden settings, or food processing factories, the potential for robots to handle granular materials is significant. Therefore, to achieve human-like precision and safety in these environments, robots must learn to manipulate these materials based on their physical properties~\cite{schenck2017learning}.

However, learning to manipulate granular materials based on physical information presents significant challenges for robots. Firstly, granular materials consist of numerous particles, resulting in a high-dimensional state space~\cite{narain2010free}. 
This complexity imposes substantial computational costs on sample-based planning and exploration-based learning methods~\cite{Softgym}. Another major challenge arises from the complex and unique physical properties of granular materials. At the microscopic scale, granular materials exhibit rich inter-particle interactions, where normal contact forces and tangential frictional forces between particles govern their motion. Due to their highly dissipative nature and macroscopic discreteness, these microscopic interactions collectively shape the macroscopic physical state of the material. When particle interactions are weak and internal friction is low, granular materials may exhibit the continuous flow characteristics of Newtonian fluids (e.g., flowing sand); when particle interactions strengthen and internal friction increases, granular materials rigidify, withstand external forces, and undergo plastic deformation, displaying solid-like properties (e.g., sand piles). Additionally, when particles are dispersed in air and interact briefly and frequently, they exhibit gaseous characteristics (e.g., dust)~\cite{Drucker-prager, hybird, gao2021simulating, semi_implicit_mpm}. These properties cause planning and learning-based manipulation to be highly expensive and intractable. 
\begin{figure}[t]
    %\vspace{0.25cm} 
    \centering
    \includegraphics[width=8.7cm]{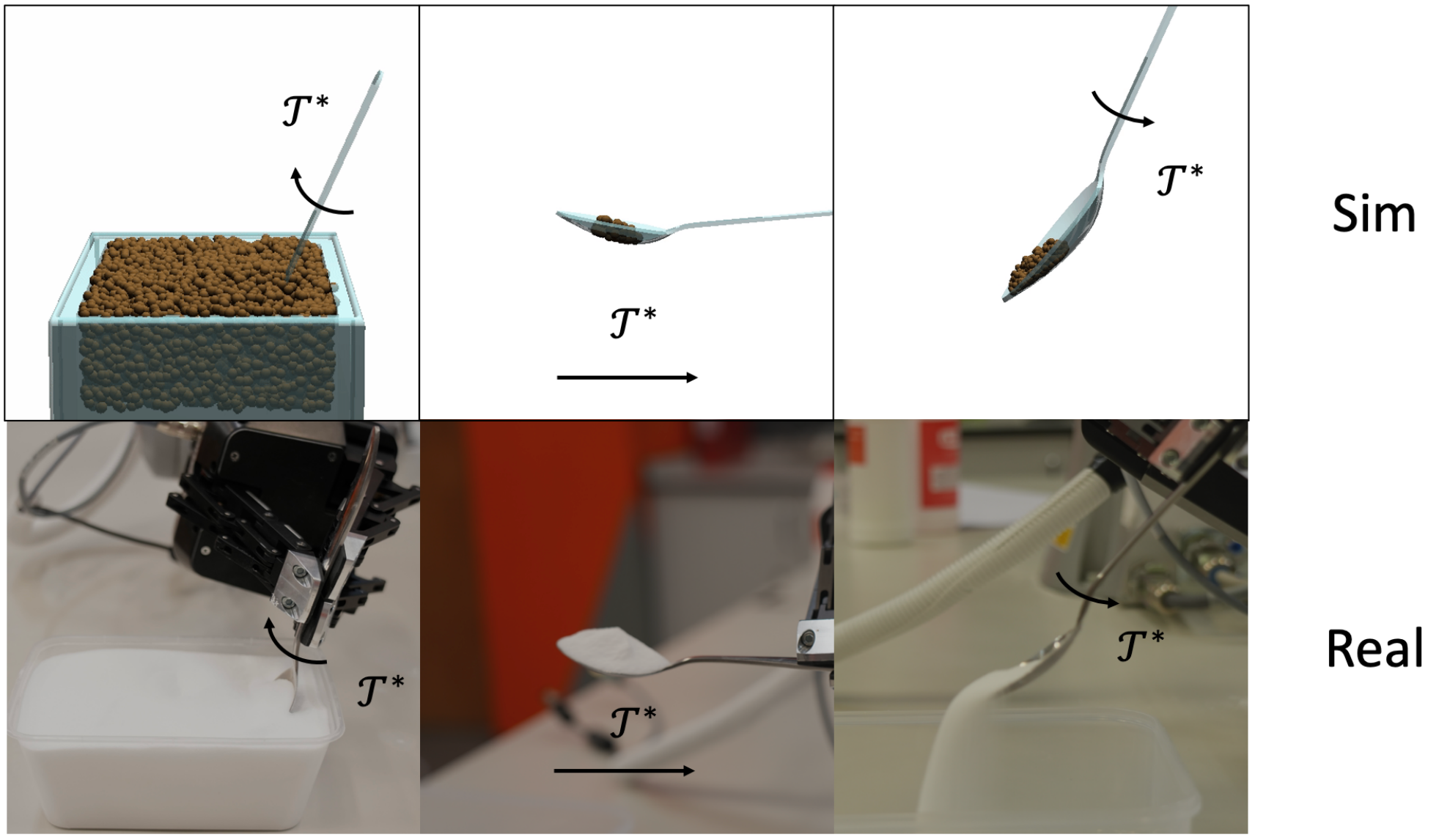}
    \caption{Granular material manipulation in our simulator (\textit{above}) and real environment (\textit{below}) for one representative task, where the agent uses a spoon to follow the optimised trajectory and completes scooping, translating, and pouring sub-tasks.}
    \label{fig:1}
    %\vspace{-0.7cm} 
\end{figure}

Due to these challenges, previous studies have predominantly relied on real-world sensor feedback~\cite{schenck2017learning,Audio,Uncertainty,kadokawa2023learning}, or deep learning (DL)-based dynamic models for state prediction, to learn granular material manipulation~\cite{GNN}, rather than using dynamic and observational physical information. These methods typically require extensive training and are computationally expensive. Despite these, they are limited to specifically-designed manipulation tasks, hindering their capabilities in handling more complex scenarios that require precise control over granular materials. In recent years, the combination of the Material Point Method (MPM) and the Drucker-Prager (DP) yield model has proven effective for large-scale numerical simulations of granular materials~\cite{Drucker-prager}. Furthermore, the use of the Taichi programming language~\cite{taichi}, with its GPU-accelerated parallel computing capabilities, automatic differentiation (AutoDiff)~\cite{difftaichi}, and efficient mesh-based operations~\cite{meshtaichi}, has made the integration of such physical simulation with learning frameworks feasible~\cite{fei2021revisiting}. Currently, simulations and learning for soft-body~\cite{plb}, fluid~\cite{fluidlab}, and elastic material property identification~\cite{DPSI} tasks have been implemented using Taichi. However, there remains a significant gap in addressing granular materials with more complex physical properties.

In this paper, we address the challenge of learning granular material manipulation in kitchen scenarios through a physics-informed approach, eliminating the need for real-world data collection or learning surrogate dynamics models, as required in previous works. To accurately capture the physical properties of such material and effectively leverage them for manipulation tasks, our framework is divided into three components: a differentiable physics-based simulator, an automatic demonstration generation module, and a demonstration-guided reinforcement learning (RL) module. Specifically, our simulator is based on the Moving Least Squares Material Point Method (MLS-MPM), incorporating the DP yield model for granular materials, and either the von Mises yield model or hydrostatic stress formulations for gradient-stable materials. This enables accurate modelling of physical behaviours during physical interaction with agents, such as robot end-effectors. The simulation is made differentiable through the AutoDiff mechanism of Taichi. Given the complex properties of granular materials, directly employing AutoDiff for gradient-based trajectory optimisation is infeasible due to gradient instability~\cite{li2023difffr}. To address this, we train an RL policy to manipulate granular materials with demonstration trajectories obtained by applying gradient-based trajectory optimisation to Newtonian fluids or elasto-plastic materials, which share similar physical properties with granular materials but produce stable gradients. These automatically generated trajectories enable rapid and effective learning for the RL agent.

We evaluate the performance of our framework by conducting common granular material transportation tasks in kitchen scenarios using a spoon, a scoop, a seasoning bottle, and a shovel. For tasks involving a spoon and a scoop, each task is divided into three sub-tasks: scoop, translate, and pour (see Fig.~\ref{fig:1}), which are trained separately. The challenge of training for long-horizon tasks is addressed by connecting the trained policies using a skill-chaining method. In addition, to minimise the gap between simulation and reality, our simulated environment is configured based on the physical parameters of real materials and the actual laboratory conditions. Experimental results show that the policies trained within our framework are capable of executing the complex task of transporting granular materials in both simulated and real-world scenarios, outperforming learning-based approaches using deep RL~\cite{sac, ppo, td3, ddpg} or Imitation Learning (IL)~\cite{bc,gail}, and other prior granular manipulation methods~\cite{GNN}. Moreover, another advantage of our framework lies in its capability of generating robust policies across diverse material properties that can be readily integrated into various RL algorithms. In summary, our key contributions are as follows:
\begin{itemize}[leftmargin=1em]
\item A flexible physics-informed robot learning framework for granular material manipulation based on differentiable simulation, allowing efficient RL-based learning through skill chaining and demonstrations. 
\item A differentiable simulator for robotic manipulation of granular material, allowing efficient granular manipulation simulation and gradient-based trajectory optimisation.
\item An automatic demonstration generation method based on differentiable simulation to replace labour-intensive human demonstrations.
\item Simulation and real-world experiments that demonstrate the superior performance achieved by the chained RL policies in long-horizon multi-step material transportation tasks.
\end{itemize}

The rest of the paper is organised as follows. Section~\ref{sec:related-work} discusses related works. Section~\ref{sec:method} describes our proposed learning framework. Section~\ref{sec:exp} presents extensive experimental results in both simulated and real-world settings. Section~\ref{sec:con} concludes the work and discusses future work.

\section{RELATED WORK}\label{sec:related-work}
Considering that granular materials are primarily represented as particles, we begin by reviewing prior work on particle manipulation in both real-world and simulation-based settings. We then extend this perspective by introducing a physics-aware control optimisation framework that bridges differentiable modelling and policy learning. Finally, we relate our approach to recent advances in granular simulation.

\subsection{Real-Environment-Based Particle Manipulation}

Manipulating particles is an active research area. An intuitive approach is to learn from real-world data. This can be in the form of learning from human demonstrations~\cite{huang2021robot}. 
Many forms of real-world sensory feedback have been used, including visual information~\cite{schenck2017learning,Uncertainty,schenck2017visual,do2019accurate,Stir_to_pour,zhang2020learning,suh2021surprising}, external physical properties~\cite{kadokawa2023learning,Stir_to_pour,Adaptable_pouring}, as well as auditory information~\cite{Audio}. For example, in~\cite{schenck2017learning}, a Convolutional Neural Network (CNN) is proposed to predict future states using height maps computed from depth images of granular materials. In addition, density is incorporated as an input for a CNN and RL methods are employed to enable the robot to gather or disperse granular materials on a surface~\cite{zhang2020learning}. Similarly, a self-supervised learning method is developed for scooping target-quality granular foods by integrating height maps and density~\cite{Uncertainty}. The proposed network also takes into account cognitive uncertainty for effective training. This vision-based feedback method has also been combined with linear models to move piles of small objects to designated target areas~\cite{suh2021surprising}. Rather than collecting feedback from RGB-D cameras, mechanical vibration information is leveraged in the form of audio produced during the manipulation of granular materials~\cite{Audio}. Using a learning framework based on CNN and Recurrent Neural Network (RNN), the robot is trained to execute shaking and dumping actions. These methods neglect the interactions between particles, which inevitably impact the results of robotic manipulation. Additionally, a more significant challenge is the excessive reliance on real-world data. Collecting data from the real-world environment and training learning models are both time-consuming processes.
\subsection{Simulation-Based Particle Manipulation}
Some recent works allow robots to learn to manipulate materials through simulations. A common approach involves using trained DL models~\cite{GNN,CConv,TIE,DPI_NET} as surrogate models to approximate their physical properties. Graph Neural Network (GNN) is a popular choice due to its ability to represent particles and the physical properties between particles as nodes and edges~\cite{DM_GNN,DM_GNN_Rigid}. In~\cite{DPI_NET}, GNN is employed to simulate material dynamics and combined with prediction and control algorithms to enable fluid manipulation. In~\cite{GNN}, GNN is used to estimate the interactions between particles and a cup. The manipulation trajectory is then optimised through a population-based optimiser. The approach of employing GNN to learn unified particle dynamics has also been demonstrated to achieve an optimal balance between efficiency and effectiveness in manipulating object piles when utilising dynamic-resolution particle representations~\cite{Dynamic_GNN}. Training such models typically requires a significant amount of data and time, and the simulation accuracy is often unsatisfactory. Moreover, any changes in the physical properties, such as friction angle, demand the regeneration of training data and retraining, imposing impracticalities. Other studies propose to learn by combining real-world data with low-resolution simulators. A likelihood-free Bayesian inference framework~\cite{Inferring} is integrated with a Discrete Element Methods (DEM)~\cite{DEM} simulator, through which input depth images are used to infer material properties, allowing the robot to better learn granular material manipulation tasks. This data-driven calibration method for DEM simulators has also been applied to granular media-related locomotion~\cite{zhu2019data}. Additionally, a low-fidelity simulator is used in~\cite{Adaptable_pouring} to simulate the physical properties of fluids and is combined with actual measurements to allow the robot to pour fluid into different containers and avoid spills. However, these learning processes still require multiple interactions between the agent and the environment.

\subsection{Physics-Aware Control Optimisation}

Recent research has investigated the integration of physical principles not only for modelling but also to guide control policy learning. A prominent example is Deep Lagrangian Networks (DeLaN)~\cite{lnn,lnnijrr}, which embed Lagrangian mechanics into neural networks to learn structured, physically consistent dynamics that produce joint torques. By introducing physics-based inductive biases, these models demonstrate improved generalisation under limited data and offer enhanced interpretability. However, DeLaN is primarily designed for rigid-body or articulated systems, where control is typically low-dimensional and focused on internal dynamics. Consequently, it lacks the capacity to model interactions with complex, deformable environments, such as the coupled dynamics between manipulators and granular media. In tasks where fine-grained contact and environmental feedback are crucial, this limitation renders DeLaN insufficient. Despite leveraging physical priors and differentiable models, it cannot handle the high-dimensional, nonlinear, interaction-rich nature of granular manipulation.

Another relevant approach is the Deep Off-Policy Iterative Learning Control~\cite{dopilc}, which uses gradients from differentiable simulators to refine control policies. By computing value-function gradients through dynamics Jacobians and reward gradients, it refines policies iteratively. However, this introduces second-order derivatives, requires smooth rewards and dynamics, and is computationally intensive and sensitive to numerical instability, especially in non-smooth or high-dimensional settings. It also requires extensive tuning and struggles with environments involving discontinuities. Moreover, its reliance on single-step rollouts limits long-term reasoning and credit assignment, making it less effective in sparse reward scenarios. Crucially, the method depends on differentiable rewards, which restricts its applicability to tasks with discrete feedback or binary success conditions. In contrast, our method avoids assumptions of reward smoothness or higher-order differentiability, enabling robust optimisation in long-horizon, contact-rich granular manipulation tasks where sparse or discontinuous rewards are often intrinsic to task success.
\subsection{Physics Simulation for Granular Materials}
As discussed in Section~\ref{sec:intro}, granular materials exhibit distinct physical properties across different scales. An early method frequently employed to simulate such materials involved treating particles as individual entities at the microscopic level to accurately model the interactions between them~\cite{Inferring,DEM,bell2005particle,cundall1980discussion,gallas1992convection}. However, tracking states and contacts for each particle poses a substantial computational burden~\cite{pancheshnyi2008numerical}. In recent years, the use of continuum models to simulate granular materials at the macroscopic level has gained favour among researchers. These continuum models do not explicitly represent individual particles, making them well-suited for simulating large quantities of particles. A prominent example is the hybrid Eulerian-Lagrangian MPM, which effectively captures high visual detail in particle dynamics at a relatively low cost. The hybrid nature of the MPM not only allows the use of Cartesian grids to efficiently handle collisions and fractures but also enables grid-based implicit integration~\cite{MPM_review}. The MPM has been shown to excel in simulating plasticity~\cite{su2023generalized,gao2017adaptive,schreck2020practical}, elasticity~\cite{su2023generalized,han2019hybrid,fang2019silly}, viscosity~\cite{su2023generalized,fang2019silly}, inelastic flow~\cite{qu2023power}, the coupling of granular materials with rigid bodies~\cite{narain2010free,takahashi2021frictionalmonolith}, and the mixing of granular materials with fluids~\cite{gao2018animating,tampubolon2017multi}. Even when compared to other Lagrangian methods that also adopt the continuum assumption, the MPM introduces stronger numerical viscosity~\cite{fei2021revisiting,yang2017unified} and simplifies the coupling of various materials. Furthermore, this method uses a continuous description of governing equations and easily incorporates user-controllable elasto-plastic constitutive models. Building on this foundation, several improved versions of the MPM have also been proposed. For example, unlike the traditional MPM using B-spline basis functions, MLS-MPM employs MLS shape functions $\Phi(\bm{x})$ as its basis functions, where $\bm{x}$ are the locations. The efficiency of MLS-MPM stems from approximating the previously computed affine velocity matrix $C^{n+1}_{p}$ as the Eulerian velocity gradient $\nabla v^{n+1}$ during the update of particle-wise deformation gradient $F_{p}$ in Lagrangian view~\cite{mls-mpm}, where $p$ and $n$ denote particle quantities and timesteps. Therefore, in this study, we follow the works in~\cite{Drucker-prager} and~\cite{mls-mpm} to model granular materials using more efficient MLS-MPM with the DP constitutive model. The simulated physical information is then utilised by the learning model within our framework.
\section{METHOD} \label{sec:method}

\begin{figure*}[t]
    %\vspace{-0.3cm} 
    \centering
    \includegraphics[width=17cm]{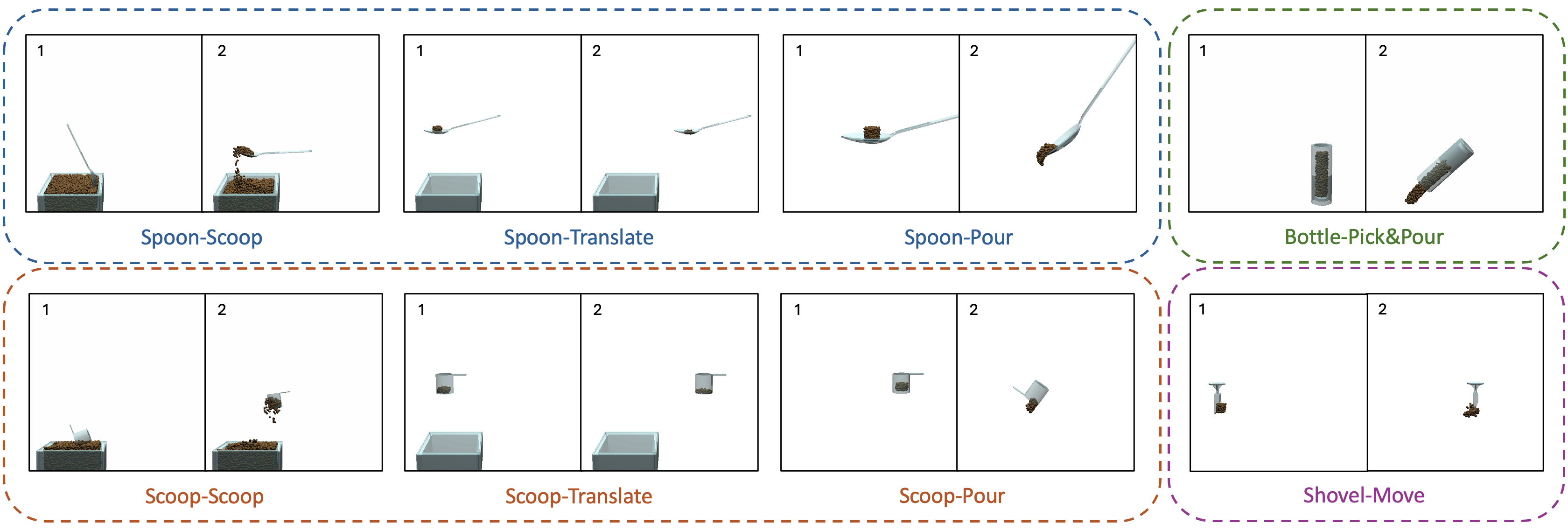}
    \caption{Illustration of the problem setting. In our study, four tasks are proposed: transporting granular materials using a spoon (blue box), a scoop (green box), a bottle (red box), and a shovel (purple box), respectively. The tasks involving the spoon and scoop consist of three sub-tasks: scooping, translating, and pouring. For each action, image 1 denotes the initial state, while image 2 denotes the state along the optimal trajectory trained by our model.}
    \label{fig:tasks}
\end{figure*}

\subsection{Learning Framework Overview} \label{sec:method:framework}

This work aims to develop a physics-based learning framework for granular material manipulation, enabling robots to move granular materials from one container to the designated target area. Our framework rigorously incorporates real-world scenarios to the greatest extent possible, to enable robots to perform granular material transportation tasks using various tools proficiently. Compared to previous studies~\cite{schenck2017learning,GNN}, our learning framework addresses more complex tasks that require longer trajectories. Specifically, we consider four kitchen tasks involving the transport of granular materials to plates, pots, or bins using various tools such as a seasoning bottle, spoon, scoop, or shovel. These tasks are challenging for RL algorithms to learn effectively due to long task horizons~\cite{yang}. Fig.~\ref{fig:tasks} visualises four tasks (eight sub-tasks) considered in this work. Transporting granular materials using scoops and spoons is complex due to the existence of multiple task stages: granular materials do not spontaneously appear on the spoon or scoop, requiring a sequence of coordinated movements. This complexity makes it challenging to generate desired trajectories in an end-to-end fashion. Our solution is to decompose tasks involving these two tools into three sub-tasks: scooping, translating, and pouring. Each sub-task is trained separately and the resultant policies are seamlessly chained to accomplish the transporting tasks.

\begin{algorithm}[!t]
    \SetAlgoLined
    \caption{Overall Process of Our Framework.}
    \label{alg:overview}
    \SetKwInOut{Input}{Input}
    \SetKwInOut{Output}{Output}
    
    \Input{number of iterations, sub-tasks, and demos $N_{IT}$, $N_{ST}$, $N_{D}$}
    \Output{optimal trajectory set $\mathcal{T}_{task}$}

    Set $\mathcal{T}_{task} = \emptyset$
    
    \For{\text{$iteration=1$} to \text{$N_{IT}$}}{
        \For{\text{$i=1$} to \text{$N_{ST}$}}{
        $E_{f} \leftarrow$ Init env with fluid config 
        
        Set $\mathcal{T}_{i,demo} = \emptyset$
        
        \For{\text{$episode=1$} to \text{$N_{D}$}}{
        $\tau_{i,demo} \leftarrow$ \text{DEMOGEN}$(E_{f})$
        
        $\mathcal{T}_{i,demo} \leftarrow \mathcal{T}_{i,demo} \cup \tau_{i,demo}$
        
        }
        $E_{g} \leftarrow$ Init env with granule config
        
        $\tau_{i} \leftarrow$ \text{DGSAC}$(E_{g}, \mathcal{T}_{i,demo})$
        
        }

        $\tau^{*} \leftarrow$ \text{CONCAT}($\tau_{1}$, $\tau_{2}$, $\ldots$, $\tau_{N_{ST}}$)
        
        $\mathcal{T}_{task} \leftarrow \mathcal{T}_{task} \cup \tau^{*}$
        
    }
    \Return{$\mathcal{T}_{task}$} 
    % \vspace{0.5em} % Adjust vertical space for the final line
\end{algorithm}

\begin{figure}[t]
    %\vspace{0.3cm} 
    \centering
    \includegraphics[width=8.7cm]{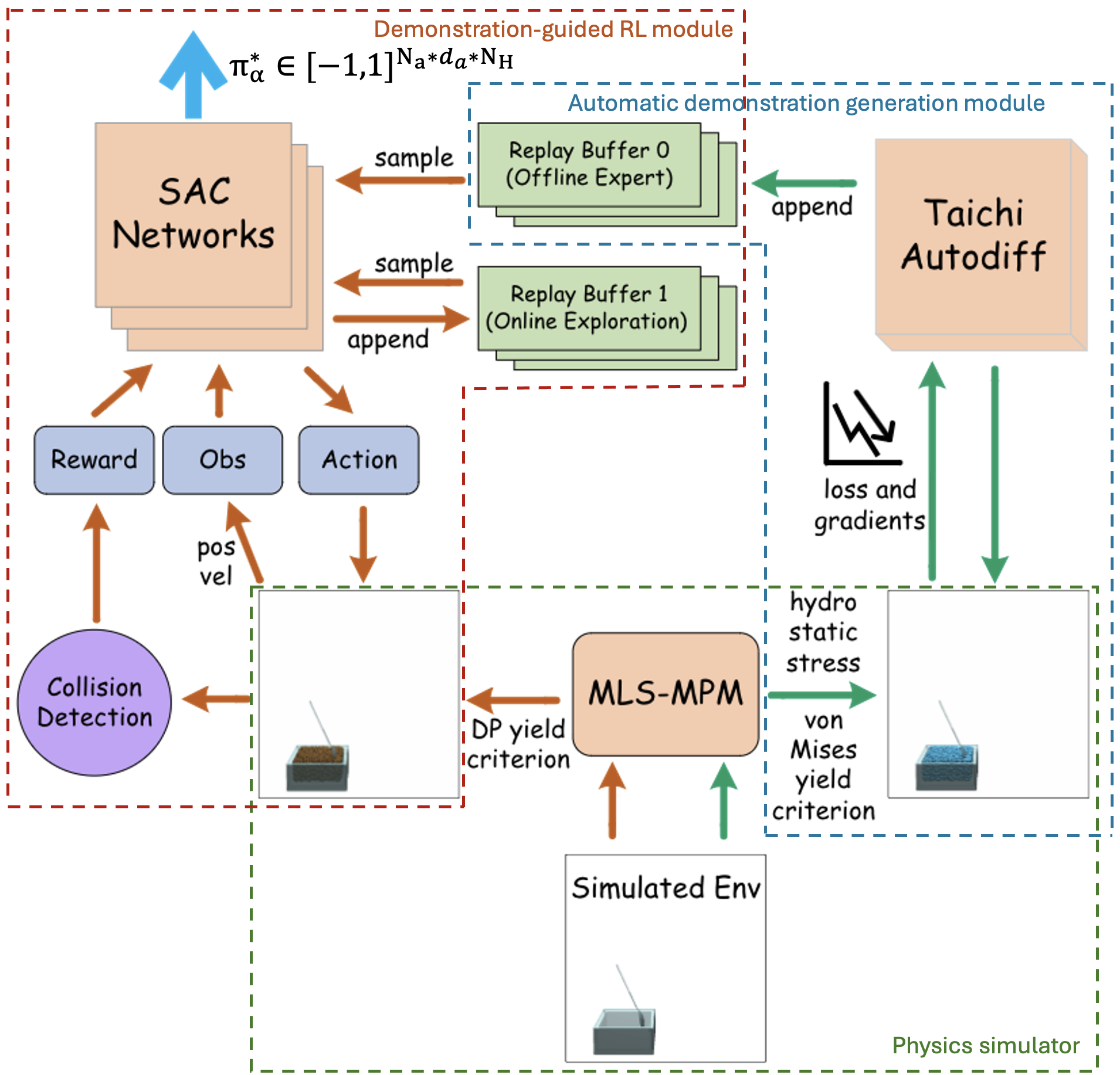}
    \caption{Workflow of the proposed learning framework. Green arrows: imperfect demonstrations generated via gradient-based optimisation with a fluid or elasto-plastic material model. Brown arrows: SAC training with dual replay buffers — a fixed buffer storing the demonstrations and an updated buffer collecting data from interaction with the actual granular dynamics. The final output is an $N_a*d_a$-dimensional policy over $N_H$ time steps.}
    \label{fig:3}
    %\vspace{-0.5cm} 
\end{figure}

As depicted in Fig.~\ref{fig:3}, the framework is comprised of three main components, including a physics simulator based on the MLS-MPM approach (Section~\ref{sec:simulator}), an automatic demonstration generation module (Section~\ref{sec:adg}), and a demonstration-guided RL module (Section~\ref{sec:dgsac}). The pseudo-code for the overall training process for each task is presented in Algorithm~\ref{alg:overview}. For each sub-task in every training iteration, we first initialise our environment with fluids ($E_{f}$ in line 4), which is then provided to the demonstration generation module, yielding $N_{D}$ fluid manipulation demonstration trajectories $\tau_{i,demo}$ (Line 7). These trajectories collectively constitute the demonstration set $\mathcal{T}_{i,demo}$ (Line 8), where $i$ denotes the respective sub-task. Subsequently, the environment is reinitialised with granular materials, denoted as $E_{g}$ (Line 10). This reconfigured environment, along with the demonstration set $\mathcal{T}_{i,demo}$, is then passed to our demonstration-guided RL module (Line 11). After training, the RL module produces an optimal trajectory for each sub-task, which is sequentially concatenated (Line 13), forming the optimal trajectory set $\mathcal{T}_{task}$ (Line 14). While our primary formulation employs fluid-based materials for demonstration generation, it can be naturally extended to gradient-stable materials for tasks where fluid modelling is inadequate, as discussed in Section~\ref{sec:exp:elastic}.
\subsection{Physics-based Simulation} \label{sec:simulator}

We employ the MLS-MPM to simulate the contact dynamics between granular materials and the agent. As with other MPM methods, MLS-MPM adopts the continuum description for the governing equations and discretises them by a collection of particles and a background Euler gird. It keeps track of positions, velocities, deformation gradients, and mass of the Lagrangian particles, but uses a fixed Eulerian grid to handle interactions and calculate forces. For more details about the MLS-MPM and related techniques, please refer to~\cite{mls-mpm}. We use the elasto-plastic continuum assumption to simulate the dynamics of granular materials. Specifically, we adopt the Fixed Corotated Constitutive Model and, following~\cite{Drucker-prager}, employ the DP yield criterion for plastic deformation projection, which is used to realistically simulate the plasticity of large-scale free-flowing granular materials. The amount of plastic deformation $y(\sigma)$ can be calculated as:
\begin{equation}
y(\sigma) = \left\Vert \sigma - \frac{tr(\sigma)}{d}I \right\Vert_{F} + \frac{d\lambda+2\mu}{2\mu}tr(\sigma)\sqrt{\frac{2}{3}}\frac{2\sin{\phi_{f}}}{3-\sin{\phi_{f}}}
\end{equation}
where $d$ represents the spatial dimension, $\phi_{f}$ denotes the user-controllable friction angle, $\lambda$ and $\mu$ are the Lam\'e constants of the material, and $\sigma$ denotes stress. If stress lies within the yield surface, i.e., $y(\sigma)\le 0$, then no plasticity occurs. Otherwise, depending on whether there is resistance to motion or dynamic friction, the deformation gradients of the particles are projected onto the tip or side of the yield surface. This projection process is integrated into the particles-to-grid stage of the MLS-MPM, thereby influencing the calculation of altering the deformation gradients of the particles and their position updates in the subsequent frames. For fluid simulation, consistent with previous work~\cite{fluidlab}, we reset the deformation gradient of the particles to a diagonal matrix at each frame according to the hydrostatic stress formula. 

Additionally, based on the work in~\cite{mpm_snow}, we construct the collision model for the particles and the rigid bodies (the tools manipulated by the agent) in our simulator, with the idealised assumption that only sliding friction exists in the tangential direction. This assumption, which overlooks a few rare scenarios, allows for more efficient training. To enable the agent to learn to avoid collisions with the static containers, in addition to the particle-agent collision detection, our learning framework also incorporates collision detection between rigid bodies, which is based on the signed distance function (SDF). We decompose the container into rigid-body particles in our simulated environments and represent the states of rigid-body particles with an $N_{r} \times d_{r}$ matrix, where $N_{r}$ is the quantity of rigid-body particles. Since they are stationary, $d_{r}$ is set to 3, including only the particle positions. Collision detection between the agent and the rigid body is achieved by monitoring the directed distance $\vec{d}$ from the particles to the tool surface at each timestep.

\subsection{Automatic Demonstration Generation Module} \label{sec:adg}

Given the complexity of particle projection onto the DP yield surface in the principal stress space, directly employing differentiation for gradient-based trajectory optimisation in granular material manipulation proves to be challenging. We adopt the concept of transfer learning by using the trajectories optimised through gradient-descent for manipulating fluids as demonstrations for learning to manipulate granular materials. As shown in Algorithm~\ref{alg:demogen}, we created an automatic demonstration generation process based on the automatic differentiation (autodiff) function provided by TaiChi, which allows automatic gradient evaluation to generate derivative functions for forward computation functions and a tape to record the order of executing these functions, with which gradients are computed by traversing the derivative functions in the backward order according to the tape recording~\cite{difftaichi}.

Before training, the trajectory is initialised to zero in all dimensions, meaning that the agent does not execute any actions. The loss and environment are first reset to their initial configurations for each optimisation iteration (Line 3 of Algorithm~\ref{alg:demogen}). At each timestep within the horizon $N_{H}$, the action is selected based on the trajectory $\tau$ (Line 5), executed in the INTERACTION (Line 7), and the loss is then calculated based on the state of the fluid particles after execution (Line 8). The INTERACTION process includes agent movement, MPM-based fluid simulation, and the handling of collisions between the agent and fluid particles (further details can be found in~\cite{fluidlab,mls-mpm,mpm_snow}). Subsequently, the gradients are reset (Line 10), and the INTERACTION is executed in reverse over $N_{H}$ steps (Line 15). The gradient of the agent's action at each step is computed (Line 17) through a single forward calculation and the back-propagation of the INTERACTION (Line 16) and loss computation (Line 14). Finally, the gradients of the entire trajectory can be used to optimise the trajectory through the reverse execution (Line 19). This AutoDiff-based backward differentiation method is highly efficient for computing gradients of complex functions.

We define the weighted Manhattan distance $\mathcal{DW}$ from particles to target positions as the primary component of the loss function for our gradient-based trajectory optimisation:
\begin{equation}
\mathcal{DW}_{\alpha}(i,j)= \mathcal{W}_{\alpha}^{x}|x_{i}-x_{j}| + \mathcal{W}_{\alpha}^{y}|y_{i}-y_{j}| + \mathcal{W}_{\alpha}^{z}|z_{i}-z_{j}|
\end{equation}
where $\mathcal{W}_{\alpha}$ represents the weights of different sub-tasks $\alpha$ along the Cartesian coordinate axes, $x_i$, $y_i$ and $z_i$ denote the Cartesian coordinates of the target position, and $x_j$, $y_j$ and $z_j$ are the Cartesian coordinates of the current particle. 
An optimisation is determined to be converged when 1) the loss value is above a sub-task-specific threshold and 2) the difference between two consecutive loss values is smaller than $5\%$ for more than $5$ iterations.

The method of automatic demonstration generation eliminates the substantial costs associated with acquiring human demonstrations. In addition, it allows the forming of a pipeline between the demonstration generation module and the demonstration-guided RL module. This streamlining facilitates an efficient learning process, allowing the RL models to benefit from automatically generated demonstrations of similar material without the need for extensive human participation. 

\begin{algorithm}[!t]
    \SetAlgoLined
    \caption{\text{DEMOGEN}$(E_{f})$.}
    \label{alg:demogen}
    \SetKwInOut{Input}{Input}
    \SetKwInOut{Output}{Output}
    \SetKwInOut{Configs}{Configs}
    
    \Input{fluid manipulation environment $E_{f}$}
    \Configs {number of optimisation iterations $N_{E}$, sub-task horizon $N_{H}$, and initial policy $\pi_{init}$}
    \Output{fluid manipulation optimal trajectory $\tau_{demo}$}

    $\tau \leftarrow \tau_{init}$
    
    \For{\text{$iteration=1$} to \text{$N_{E}$}}{

        Reset $loss$ and Env $E_{f}$
    
        \For{\text{$step=1$} to \text{$N_{H}$}}{
        
        $action \leftarrow \tau[step]$

        Save $E_{f}$ states at this $step$
        
        Execute one step \text{INTERACTION}$(action, E_{f})$

        Compute $loss[step]$

        }

        Reset $grads$
        
        \For{\text{$step=N_{H}$} to \text{$1$}}{

        Load $E_{f}$ states at this $step$

        $action \leftarrow \tau[step]$

        Back-propagate the computation of $loss[step]$

        Execute one step \text{INTERACTION}$(action, E_{f})$

        Back-propagate \text{INTERACTION}$(action, E_{f})$ 

        $grads[step] \leftarrow$ the gradient of agent action

        }
   
    $\tau \leftarrow$ \text{ADAM}$(grads)$
        
    }
    $\tau_{demo} \leftarrow \tau$ 
    
    \Return{$\tau_{demo}$} 
    % \vspace{0.5em} % Adjust vertical space for the final line
\end{algorithm}

\subsection{Demonstration-guided RL Module}\label{sec:dgsac} 

The RL module is based on the off-policy Soft Actor-Critic (SAC) algorithm, which optimises a stochastic policy and a soft Q function with an extra entropy maximisation term in the learning objective~\cite{sac}. The training process for the RL module is presented in Algorithm~\ref{alg:dgsac}. We introduce an additional replay buffer specifically for storing demonstration data in the original SAC algorithm. This modification addresses the issue in the original SAC model where demonstration data would be discarded from the replay buffer as experiences populate the buffer in a first-in-first-out manner. Before the optimisation iterations begin, the demonstration trajectories are executed first (Line 7). At each step $t$, the state $S_{t}$, action $A_{t}$, reward $R_{t}$, and next state $S_{t+1}$ are added to an additional replay buffer $RB_{demo}$ (Line 8). So during the optimisation iterations, by sampling from both replay buffers for training (Line 19), our model can not only learn directly from expert demonstrations but also maintain the ability to self-explore the environment. Consistent with the original SAC method, we save the policy with the highest average reward during evaluation as the optimal policy. In our model, demonstration data influence network weights as soon as they are added to the replay buffer (Line 9), thereby accelerating the initialisation phase of the learning process. 

When calculating the reward at each step, we incorporate an elite particle selection process (Line 6 and Line 16). We define the elite particle set $\Upsilon$ as:
\begin{equation}
\label{eq1}
\Upsilon=sort(O_{p},\mathcal{DW}_{s}(i,goal))[:\widetilde{N}_{p}^{*}]
\end{equation}
where $\widetilde{N}_{p}^{*}$ is the number of elite particles, $O_{p}$ denotes the particle observation state space for a single sub-task, and $sort$ represents the function of sorting particles based on the $\mathcal{DW}$, where the first $\widetilde{N}_{p}^{*}$ particles are selected. This implies that only $\widetilde{N}_{p}^{*}$ particles closest to the target position are utilised for calculating the reward at this step. This selection method not only enhances computational efficiency and optimises resource allocation but, more importantly, improves the stability and quality of samples during training. In certain scenarios, the states of some particles are less relevant to the training process. This is particularly evident in the scooping sub-task, where the majority of particles remain stationary in the container throughout the time horizon. By calculating rewards based solely on the states of particles that are highly relevant to the task objectives, the model can more accurately learn the desired behaviours and improve its generalisation capabilities.

\begin{algorithm}[!t]
    \SetAlgoLined
    \caption{\text{DGSAC}$(E_{g}, \mathcal{T}_{demo})$.}
    \label{alg:dgsac}
    \SetKwInOut{Input}{Input}
    \SetKwInOut{Output}{Output}
    \SetKwInOut{Configs}{Configs}
    
    \Input{granular manipulation environment $E_{g}$, and demonstration set $\mathcal{T}_{demo}$}
    \Configs {number of optimisation iterations $N_{E}$, sub-task horizon $N_{H}$}
    \Output{granular manipulation optimal trajectory $\tau^{*}$}

    Init RL model, clear replay buffers
    
    \For{$\tau_{demo}$ in $\mathcal{T}_{demo}$}{

    $S_{1} \leftarrow$ Reset Env $E_{g}$
    
    \For{\text{$t=1$} to \text{$N_{H}$}}{
        
    $A_{t} \leftarrow \tau_{demo}[t]$

    Elite particle selection
        
    $R_{t}, S_{t+1} \leftarrow$ \text{INTERACTION}$(A_{t}, E_{g})$

    $RB_{demo} \leftarrow RB_{demo} \cup \{S_{t}, A_{t}, R_{t}, S_{t+1}\}$

    Sample from $RB_{demo}$, update networks

    }        
    }
    
    \For{\text{$iteration=1$} to \text{$N_{E}$}}{

        $S_{1} \leftarrow$ Reset Env $E_{g}$
    
        \For{\text{$t=1$} to \text{$N_{H}$}}{
        
        $A_{t} \leftarrow explore(S_{t})$

        Elite particle selection
        
        $R_{t}, S_{t+1} \leftarrow$ \text{INTERACTION}$(A_{t}, E_{g})$
        
        $RB_{exp} \leftarrow RB_{exp} \cup \{S_{t}, A_{t}, R_{t}, S_{t+1}\}$

        Sample from $RB_{demo}$ and $RB_{exp}$, update networks

        }
    Evaluate at intervals. Save the policy as the best $\tau^{*}$ if the average training reward is the highest so far
        
    }
    
    \Return{$\tau^{*}$} 
    % \vspace{0.5em} % Adjust vertical space for the final line
\end{algorithm}

\subsection{Skill Chaining}

\begin{figure*}[t]
    %\vspace{-0.3cm} 
    \centering
    \includegraphics[width=17cm]{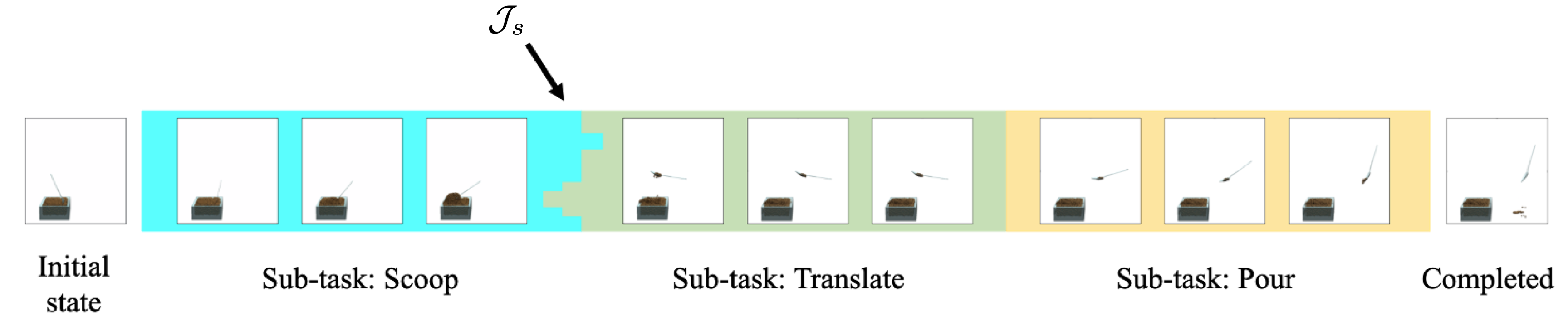}
    \caption{We employ the concept of skill chaining, innovatively integrating an Euler angle objective function $\mathcal{J}_{s}$ within the learning paradigm of the scooping sub-tasks. This function is designed to drive the agent towards achieving a seamless connection between scooping and translating actions.}
    \label{fig:2}
    %\vspace{-0.3cm} 
\end{figure*}

Drawing from the concept of skill chaining, an Euler angle objective function $\mathcal{J}$ is introduced to ensure a smoother transition between sub-tasks. This function incentivises the agent to have an appropriate posture after scooping, thus creating a chain. Specifically, it is desirable for the robot to scoop the material and ensure the tool is relatively level before transitioning to the translation motion, as illustrated in Figure~\ref{fig:2}. This is achieved by extracting the 4D quaternion rotation vector ${\bf{q}}=[q_{w},q_{x},q_{y},q_{z}]$ of the agent at the end of the sub-task of the duration of $T_{s}$. The extracted vector is initially transformed into Euler angles $\vartheta$, constituting a set of Euler angles denoted as $\Theta$:  
\begin{equation}
\left[
\begin{array}{c}
   \vartheta^{x} \\
   \vartheta^{y} \\
   \vartheta^{z}
\end{array}
\right]
=
\left[
\begin{array}{c}
    \tan^{-1}(2(q_{w}q_{x}+q_{y}q_{z}), 1-2(q_{x}^{2}+q_{y}^{2})) \\
    \sin^{-1}(2(q_{w}q_{y}-q_{x}q_{z})) \\
    \tan^{-1}(2(q_{w}q_{z}+q_{x}q_{y}), 1-2(q_{y}^{2}+q_{z}^{2}))
\end{array}
\right]
\end{equation}where $\vartheta$ represents Euler angles, and they form a set $\Theta$. The Euler angle objective function for each sub-task $\alpha$ is formulated as:
\begin{equation}
\mathcal{J}_{\alpha}=\beta_{\alpha}^{ea}(\gamma_{\alpha}^{ea} - \sum_{j=0}^{N_{a}}\sum_{\vartheta \in \Theta}\mathcal{C}_{r}(\varepsilon_{\vartheta})|\vartheta_{T_{s}-1}(j)-\vartheta_{goal}(j)|)
\end{equation}
where $\varepsilon$ and $\mathcal{C}_{r}$ denote the adjustable rotation control vector and function to govern the rotational degrees of freedom across distinct sub-tasks, $\beta$ and $\gamma$ are constants. In the scooping sub-tasks, we set a higher $\beta$ value for the Spoon-Scoop sub-task than the Scoop-Scoop sub-task. This design choice was deliberate, as the transition process in the spoon-based task is more prone to material loss, thereby necessitating a higher level of precision in skill transitions.

\subsection{Problem Formulation} \label{pf}

Each sub-task in our work can be represented as a separate Markov Decision Process (MDP). Briefly, an MDP can be represented as a four-element tuple: $(s_{t}, a_{t}, p(s_{t+1}|s_{t}, a_{t}),$ $r(s_{t}, a_{t}))$, where $s_{t}$ and $a_{t}$ denote the system state and the action at timestep $t$, respectively. $p(s_{t+1}|s_{t}, a_{t})$ is the transition probability function for reaching the next state $s_{t+1}$ under the state $s_{t}$ and action $a_{t}$. $r(s_{t}, a_{t})$ is the reward obtained after the state transition. The following subsections introduce the details of state/observation representation, action space, and rewards.

%\textit{1)} 
\textbf{\textit{States:}} We define the state for each sub-task with two main components, including the states of the granular particles and the agent — that is, the end effector of the manipulator. A particle state matrix of size $N_{p} \times d_{p}$ is employed to represent the state of the particles, where $N_{p}$ is the total number of particles in the system and $d_{p}$ is the dimensionality of state for each particle. In our framework, $d_{p}$ is set to 6, representing the position and velocity of a particle in the 3D Cartesian coordinates. Furthermore, a matrix of size $N_{a} \times d_{e}$ is employed to encapsulate the state of the end effector. Here, $N_{a}$ represents the number of agents, and $d_{e}$ signifies the state dimension for each manipulator. 
In our case, there is one manipulator, hence $N_{a}=1$. 
In general, $d_{e}=7$, containing a 3D positional vector coupled with a 4D quaternion rotation vector, except for the translation and move sub-tasks, where $d_{e}$ is set to 3, as rotation is not needed in this case.

\textbf{\textit{Observations:}} 
Optimising the observational input is essential for the learning models. Providing the model with state information on all particles may escalate complexity, thereby impeding the learning process. To address this issue, we introduce a parameter denoted by $\delta_{d}$, which serves as a tunable step size, facilitating systematic down-sampling of the granular particles in environments with a large number of particles (Spoon-Scoop, Scoop-Scoop, and Bottle-Pick\&Pour). Thus, under these conditions, the number of elements observed by the agent is:

\begin{equation}
N_{o}=\lfloor \frac{N_{p}}{\delta_{d}} \rfloor d_{p} + N_{a}d_{e}
\end{equation}

%\textit{3)} 
\textbf{\textit{Actions:}} The agent in our work is capable of performing both linear translations and rotational actions. The actions are represented in a matrix of dimensions $N_{a} \times d_{a}$, where $d_{a}$ signifies the dimensionality of action-related information. Beyond linear velocities along the Cartesian coordinate axes, the control of rotational dynamics is achieved through the update of angular velocities at the three Euler angles. Furthermore, in the RL models and environments, we impose boundaries $A_{min}, A_{max} \in \mathbb{R}^{d_{a}}$, on the selection of actions, contributing to enhancing system stability and efficiency to facilitate training.

%\textit{4)} 
\textbf{\textit{Rewards:}} Our eight sub-tasks are categorised into three types based on the actions involved: scooping, translating (including Shovel-Move), and pouring (including Bottle-Pick\&Pour). Each type of sub-task $\alpha$ in our framework is equipped with a unique reward function composed of multiple sub-rewards. Sub-tasks of the same type share the same reward function structure but differ in their parameters. Within the three types, a distance-centric reward is integrated to serve as the principal incentive for the acquisition of granular material manipulation skills. This is implemented by calculating the weighted Manhattan distance $\mathcal{DW}$ between the manipulated particles and their target positions at each timestep $t$:
\begin{equation}
\mathcal{R}_{\alpha}^{dist}(t) = \beta_{\alpha}^{dist}(\gamma_{\alpha}^{dist}-\sum_{i=0}^{N_{p}^{*}} \mathcal{DW}_{\alpha}(\mathcal{P}_{i}^{t},\mathcal{P}_{goal}))
\end{equation}
where $\mathcal{P}_{i}^{t}$ denotes the Cartesian coordinates of particle $i$ at the timestep $t$, while $N_{p}^{*}$ represents the number of particles in the particle observation state space $O_{p}$ for a single sub-task. $\beta$ and $\gamma$ are constants representing weights and biases, respectively. 

In the pouring-related sub-tasks, we introduce two sparse rewards to encourage the agent to pour out particles and accurately deposit them into the designated area. The reward function at timestep $t$ for these sub-tasks is:
\begin{equation}
\label{rp}
\begin{split}
\mathcal{R}_{p}(t)=\mathcal{R}_{p}^{dist}(t)+\beta_{p}^{p}\Delta(T_{s}-1)\mathbbm{1}^+(\mathcal{P}_{i} \in \Phi) \\
+\beta_{p}^{t} \sum_{i=0}^{N_{p}^{*}} \mathbbm{1}^+(\mathcal{P}_{i}^{'} \in \Omega_{p} | \mathcal{P}_{i} \in \Phi, \mathcal{P}_{i}^{'} \notin \Phi)
\end{split}
\end{equation}
where $\Delta(T_{s}-1)$ represents the Kronecker delta function $\delta(t-T_{s}+1)$, $\mathbbm{1}^+: X \to \{-1,1\}$ is our defined indicator function, $\Omega$ refers to the set of target positions, $\Phi$ signifies the set of positions outside the environmental boundaries, and ${P}_{i}^{'}$ denotes the position of particle $i$ at the last timestep. 

In the translating-related sub-tasks, two sparse rewards are introduced to mitigate the transportation loss during the process and to encourage the transport of particles to the target area. The reward function of these sub-tasks is given by:
\begin{equation}
\label{rt}
\begin{split}
\mathcal{R}_{t}(t)=\mathcal{R}_{t}^{dist}(t)+\beta_{t}^{n} \sum_{i=0}^{N_{p}^{*}} \mathbbm{1}(\mathcal{P}_{i} \in \Phi | \mathcal{P}_{i}^{'} \notin \Phi) \\
+\beta_{t}^{t} \sum_{i=0}^{N_{p}^{*}} \Delta(T_{s}-1) \mathbbm{1}^+(\mathcal{P}_{i} \in \Omega_{t})
\end{split}
\end{equation}
where $\mathbbm{1}$ denotes the regular indicator function. At each timestep in the scooping-related sub-tasks, only the carefully selected $\widetilde{N}_{p}^{*}$ elite particles (refer to Section \ref{sec:dgsac}) are utilised for calculating $\mathcal{R}_{s}^{dist}$. In order to avoid the results of the RL model optimisation falling into a local optimum, i.e., the agent prefers to choose not to perform rotational actions to avoid collisions, a reward is added to encourage the agent to interact with the particles. Besides the two sparse rewards in $\mathcal{R}_{t}$, another negative sparse reward is introduced to prevent collisions with the container. The reward function for scooping is defined as:
\begin{equation}
\label{rs}
\begin{split}
\mathcal{R}_{s}(t)=\mathcal{R}_{s}^{dist}(t)+\Delta(T_{s}-1)(\beta_{s}^{t} \sum_{i=0}^{N_{p}^{*}} \mathbbm{1}^+(\mathcal{P}_{i} \in \Omega_{s})+\mathcal{J}_{s})\\
+\beta_{s}^{c}\xi_{t}^{p}+\beta_{s}^{n} \sum_{i=0}^{N_{p}^{*}} \mathbbm{1}(\mathcal{P}_{i} \in \Phi | \mathcal{P}_{i}^{'} \notin \Phi) + \beta_{s}^{i} \mathbbm{1} (\xi_{t}^{r}>0)
\end{split}
\end{equation}
%\mathfrak{I}

where $\xi_{t}^{r}$ and $\xi_{t}^{p}$ denote the number of rigid-body particles and particles that collide with the agent at timestep $t$. Within all the reward functions, $\beta$ and $\gamma$ are constants representing weights and biases, respectively, and among them, $\beta^{n}$ and $\beta^{i}$ are negative values.

\section{EXPERIMENTS AND RESULTS} \label{sec:exp}

This section evaluates and analyses the experiment results of granular material manipulation tasks performed using our learning framework in simulated and real-world environments. Section~\ref{sec:exp:setup} describes the experiment setup. Section~\ref{sec:exp:whyDGSAC} demonstrates the necessity of our demonstration-guided learning method. Section~\ref{sec:exp:baseline} compares our method with popular deep RL and IL methods. Section~\ref{sec:exp:gnn} further compares our method with state-of-the-art granular manipulation approaches. Section~\ref{sec:exp:ablation} presents the ablation studies and Section~\ref{sec:exp:chaining} verifies the necessity of skill chaining. Section~\ref{sec:exp:elastic} extends demonstration generation to tasks unsuited for fluid modelling, followed by robustness tests across varying material properties in section~\ref{sec:exp:material}. The architectural flexibility of our framework is investigated in section~\ref{sec:exp:pointnet}. Finally, Section~\ref{sec:exp:real} presents the real-world experiments.

\subsection{Experiment Setup} \label{sec:exp:setup}

We simulate all sub-tasks in a $1 \times 1 \times 1$\,m workspace (Fig.~\ref{fig:2}), using 3D meshes to represent end-effector tools and containers. For collision-sensitive tasks, the container is filled with rigid-body particles to support collision detection with the agent. The simulation states are initialised based on the real-world configurations, including the positions and orientations of the agent (manipulator) and container, and the spatial distribution of granular material. Minor adjustments to the corresponding simulation parameters are sufficient to reflect the variations observed in the real environment. To improve efficiency, we simulate only task-relevant particles (fewer than 1{,}000) within the tool during translating and pouring actions. To reduce the sim-to-real gap, particle dynamics are parameterised using real-world physical properties. This setup enables the simulation to approximate real-world conditions with sufficient fidelity, allowing policies to be trained entirely in simulation and deployed without additional adaptation, which is validated in Section~\ref{sec:exp:real}. For granular materials, we set the friction angle $\phi_f = 30^\circ$, shear modulus $\mu = 416.67$, and Lamé constant $\lambda = 277.78$, corresponding to Young’s modulus $E = 1000$ and Poisson’s ratio $\nu = 0.2$, via: $\mu = \frac{E}{2(1+\nu)}$ and $\lambda = \frac{E\nu}{(1+\nu)(1 - 2\nu)}$. For fluids, we use $\mu = 0$, allowing deformation under any non-zero shear stress, with $\lambda = 277.78$. Each sub-task runs for 1{,}000 simulation timesteps.

We conducted five training runs for each sub-task, using each model separately. Considering the fluctuations in the reward values, the average reward over the last 10 episodes before the end of the training is selected as the training reward. In addition to the accumulated training reward per episode, we defined a task completion score (TCS) to assess task completion, which excludes the terms in the reward definitions that are not directly related to task completion. For all sub-tasks, the TCS incorporates a sparse reward to evaluate whether the granular material has been successfully transported to the designated area (either poured into, translated into, scooped into, or moved into). Furthermore, for the pouring sub-task, the TCS includes an additional reward that assesses whether the material is successfully poured out, while for the translation sub-task, it incorporates a penalty assessing material loss during the transportation process. For the scooping action, the TCS also includes the Euler angle objective function $\mathcal{J}_{s}$, as well as penalties for granular material loss and collisions. Overall, the TCSs assess whether the particles are poured into a specified region, whether the particles are translated to a specified position without spillage, and whether sufficient amounts of particles are scooped up without colliding with the container and within the boundary.

\begin{table}[h]
\centering
\caption{Demonstration-guided SAC Parameters.}
\label{sac}
\resizebox{0.5\columnwidth}{!}{
\begin{tabular}{c|ccc}
\toprule[1.25pt]
    gamma                & 0.99  \\
    policy learning rate            & 0.003 \\
    entropy learning rate           & 0.003 \\
    batch size           & 128   \\     
    replay buffer 0 size & 5e4   \\
    replay buffer 1 size & 1e5   \\
    hidden layers        & 2     \\
    layer size & 256 \\
    \bottomrule[1.25pt]
\end{tabular} }
\label{4}
%\vspace{-0.6cm}
\end{table}

Table~\ref{sac} summarises the hyperparameters of our DG-SAC agent. For the demonstration generation module, we employ the Adam optimiser for gradient-based trajectory optimisation, with a learning rate of 0.0001. 

All experiments are conducted on a desktop with an Nvidia RTX 3080 GPU and an Intel i7-12700K CPU. Each sub-task typically requires only 50–150 episodes for both demonstration generation and RL training, allowing our method to produce results within a relatively short time period. Table~\ref{table:time} reports the average per-episode runtime as well as cumulative runtimes for 50 and 150 episodes, covering both stages. We omit the initial overhead from Taichi’s just-in-time (JIT) compilation (first episode only) and report timings after execution stabilises.

\begin{table}[h]
\centering
\caption{Average Runtime for Demonstration and Training}
\resizebox{1.0\columnwidth}{!}{
\begin{tabular}{c|ccc|ccc}
\toprule[1.25pt]
\multirow{2}{*}{Task} &  \multicolumn{3}{c|}{Demo Generation} & \multicolumn{3}{c}{RL Training} \\ 

& 1 ep & 50 ep & 150 ep & 1 ep & 50 ep & 150 ep \\

\midrule[0.75pt]
Spoon-Scoop & 32.07s & 26min44s & 80min10s & 23.67s & 19min44s & 59min11s  \\
Spoon-Translate & 17.00s & 14min10s & 42min30s & 13.05s & 10min52s & 32min38s  \\
Spoon-Pour & 15.16s & 12min38s & 37min54s & 13.90s & 11min35s & 34min45s \\
Scoop-Scoop & 34.67s & 28min54s & 86min40s & 24.66s & 20min33s & 61min39s \\
Scoop-Translate & 15.78s & 13min9s & 39min27s & 13.19s & 10min60s & 32min58s \\
Scoop-Pour & 15.18s & 12min39s & 37min57s & 13.10s  & 10min55s & 32min45s \\
Bottle-Pick\&Pour & 15.48s & 12min54s & 38min42s &14.53s & 12min7s & 36min21s \\
Shovel-Move & 12.95s & 10min48s & 32min24s &11.62s & 9min41s & 29min3s \\
\bottomrule[1.25pt]
\end{tabular} 
}
\label{table:time}
\end{table}

\subsection{Why combining non-granular demonstrations and RL}\label{sec:exp:whyDGSAC}
\begin{figure}[t]
    \centering
    \includegraphics[width=8.7cm]{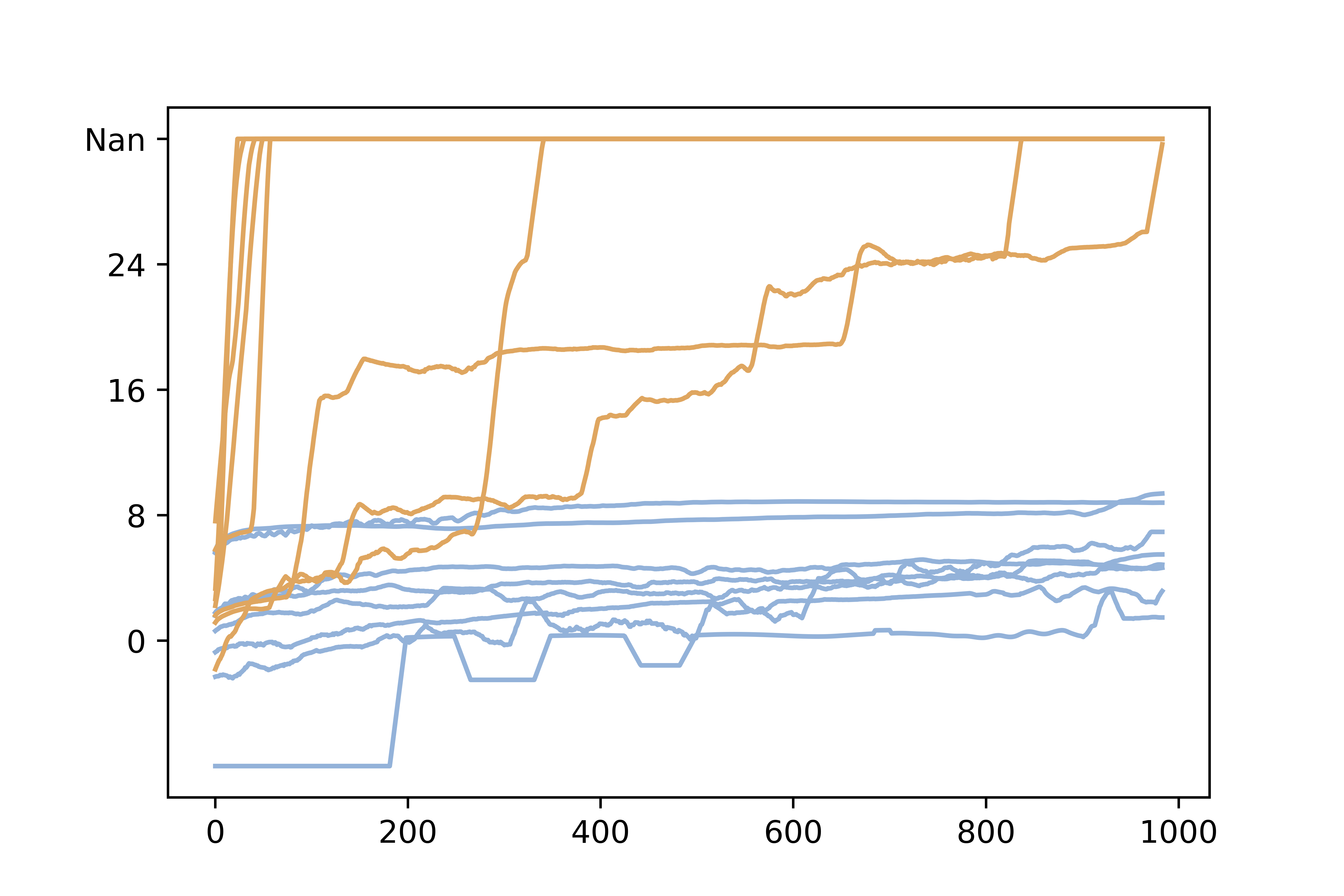}
    \caption{The variation in gradients at different timesteps during the backpropagation phase for fluids or elasto-plastic materials (\textit{blue}) and granular materials (\textit{orange}) in the first iteration of trajectory optimisation across different sub-tasks.}
    \label{fig:instable}
\end{figure}
\textbf{Exploding/unstable gradients in granular material simulation.} 
We start our experiments by examining the scale of the gradients of actions from the last timestep to the first timestep.
Fig.~\ref{fig:instable} shows the gradients (log-scale, in reversed timesteps) of actions at different timesteps with respect to the reward function for our eight sub-tasks, under both fluid (or elasto-plastic materials) and granular material configurations. ``Nan" indicates that the gradients have exceeded the maximum limit. As shown, the gradients of the manipulation actions for granular materials (orange lines) exhibit significant instability compared to those for fluids (blue lines), rapidly increasing and resulting in explosion. This is consistent with the discussions in~\cite{li2023difffr}, which highlight that the backward differentiation mode of Taichi AutoDiff does not effectively handle complex control flows, particularly those with nested loops and branching conditions. Since automatic differentiation requires unrolling the entire computational graph, it not only leads to inefficiencies in gradient computation but also exacerbates gradient explosion issues. Consequently, direct trajectory optimisation becomes impractical due to gradient instability. Therefore, RL is used as an alternative as it requires no precise gradient computation. 

\textbf{Poor generalisation of fluid-based manipulation trajectory.} Secondly, we examine the performance of the manipulation trajectory optimised with fluid simulation. In Table~\ref{table:base}, the rows starting with `Demo' present the performance of the demonstrations generated in fluid-based simulation and the rows with `DG-SAC' present our DG-SAC agent. 

For the translating sub-tasks, whether by a spoon or a scoop, both the demonstration and the trained DG-SAC agent can successfully translate the granular materials to the designated location. 
This is shown by Table~\ref{table:base} where, in all five trials of the translating sub-tasks, the TCS values reached the highest achievable score ($41.00$ and $44.10$). The specific TCS value is determined by the $\beta$ set for the sub-task and the number of granular particles $N_{p}^{*}$ involved. This is not surprising because, compared to granular materials, fluid is more likely to spill during translation. In other words, a trajectory that successfully translates fluids should be more careful and likely to transport granular materials too. 

However, for the pouring tasks, the trajectories used for pouring fluids are unsuitable for pouring granular materials. This is because the friction between the granular materials and the tool needs to be overcome by the agent applying a greater angular velocity to rotate the tool so that gravity can pull down these granular materials. For the scooping sub-tasks, as we exclude collision detection-related losses in the demonstration trajectory optimisation, the demonstration trajectories tend to collide with the container, resulting in worse performances than DG-SAC. 

These demonstration trajectories perform well in cases where the key differences in physical properties between granular materials and fluids have a limited impact on how the material reacts to the manipulation motion (external forces), such as in the translation sub-task. In these scenarios, it's more likely to find a motion that works for all materials. However, as discussed, this is not the case for general granular material manipulation, where the unique physical interactions of granular systems play a significant role in how a task should be approached. This again underscores the necessity of an RL framework that improves upon the demonstrations.

\begin{table*}[]
%\vspace{0.3cm}
\centering
\caption{The training reward (top) and task completion score (bottom) with their standard deviations for each evaluation method.}
\resizebox{1.8\columnwidth}{!}{
\begin{tabular}{c|ccc|ccc|c|c}
\toprule[1.25pt]
Task     & Spoon-Scoop & Spoon-Translate & Spoon-Pour & Scoop-Scoop & Scoop-Translate & Scoop-Pour & Bottle-Pick\&Pour & Shovel-Move  \\ \midrule[0.75pt]
PPO      & -47.23$\pm$0.99  & -75.04$\pm$4.49 & \textbf{164.55$\pm$41.13}& -65.84$\pm$16.04 &  -59.48$\pm$28.42 & 26.75$\pm$73.25 &  111.53$\pm$25.88 & -41.39$\pm$46.85 \\
SAC      & -20.39$\pm$19.69   & -72.66$\pm$0.09       & 138.23$\pm$57.69 & -33.10$\pm$39.52 &  -63.39$\pm$20.08 & 63.27$\pm$33.82 & 105.36$\pm$17.20 & 2.00$\pm$73.70 \\ 
TD3      &  -11.79$\pm$0.09 & -46.97$\pm$22.14      &  91.70$\pm$0.74 & -53.02$\pm$42.54 &  -65.80$\pm$28.81 & -14.33$\pm$0.00 & 145.60$\pm$0.01 & -2.43$\pm$76.01 \\
DDPG      & -135.67$\pm$0.15  & -74.94$\pm$0.01       &  138.81$\pm$52.29 & -0.78$\pm$0.03 &  -86.17$\pm$0.00 &  \textbf{101.08$\pm$0.17} &  \textbf{148.15$\pm$1.80} & 6.96$\pm$85.11 \\ \midrule[0.75pt]
BC  & \textbf{103.36$\pm$6.07} & 56.32$\pm$1.74 & -132.68$\pm$1.53 & 16.47$\pm$3.45 & 76.56$\pm$1.03 & -42.20$\pm$2.22  & 52.35$\pm$2.98 & -22.77$\pm$2.48 \\ 
GAIL & -146.43$\pm$64.20 & 43.21$\pm$33.40 & -96.45$\pm$4.46 & -220.84$\pm$1.41 & \textbf{88.14$\pm$0.54} & -58.94$\pm$5.85  & 62.43$\pm$116.83 & 67.92$\pm$0.01 \\ \midrule[0.75pt]
GNN~\cite{GNN}  & -188.77$\pm$56.81 & -57.59$\pm$1.26 & -162.91$\pm$5.22 & -23.50$\pm$28.30 & -65.62$\pm$6.37 & -37.68$\pm$1.43 & 56.35$\pm$21.69 & -85.00$\pm$2.23 \\ \midrule[0.75pt]
Demo   & 65.78$\pm$5.39   & 57.31$\pm$1.79 & -118.66$\pm$0.80  & 14.76$\pm$0.43 & 80.07$\pm$0.26& -42.82$\pm$0.36 & 50.33$\pm$2.83 & 15.92$\pm$0.00 \\ 
DG-SAC (Ours)   & 49.87$\pm$20.95   & \textbf{60.01$\pm$0.05}       & 49.93$\pm$6.32 & \textbf{93.13$\pm$35.39} & 86.96$\pm$4.69 &  63.91$\pm$7.20 &  98.79$\pm$0.00 & \textbf{74.75$\pm$2.66} \\ \bottomrule[1.25pt]
\toprule[1.25pt]
Task     & Spoon-Scoop & Spoon-Translate & Spoon-Pour & Scoop-Scoop & Scoop-Translate & Scoop-Pour & Bottle-Pick\&Pour & Shovel-Move  \\  \midrule[0.75pt]
PPO      & -25.37$\pm$1.10   & -41.00$\pm$0.00       & 11.59$\pm$1.07 & -57.66$\pm$16.58 &  -39.72$\pm$4.38 &  -8.57$\pm$19.93 & 14.56$\pm$5.96 & -20.98$\pm$25.40 \\
SAC      & -10.33$\pm$1.20   & -41.00$\pm$0.00       & 13.90$\pm$5.01 & -49.16$\pm$5.79 &  -36.82$\pm$10.30 & 22.69$\pm$5.02 &  22.19$\pm$5.02 & 3.28$\pm$36.81 \\ 
TD3       & 7.48$\pm$0.02  & -39.91$\pm$0.81       & 8.13$\pm$0.01 &  -45.26$\pm$38.32 &  -50.90$\pm$9.62 & -39.60$\pm$0.00 & 6.64$\pm$0.00 & 4.00$\pm$38.14 \\
DDPG      & -115.93$\pm$0.09  & -41.00$\pm$0.00       &  5.61$\pm$3.52 & 7.43$\pm$0.00 &   -44.10$\pm$0.00 & 43.79$\pm$0.16 & 6.60$\pm$0.04 & 8.34$\pm$40.86 \\ \midrule[0.75pt]
BC  & 20.75$\pm$1.63 & 39.50$\pm$1.50 & -30.45$\pm$0.00 & 12.01$\pm$1.21 & \textbf{44.10$\pm$0.00} & -39.60$\pm$0.00 & 24.74$\pm$0.16 & 1.67$\pm$1.86 \\ 
GAIL & -160.80$\pm$53.25 & 26.60$\pm$27.32 & -30.45$\pm$0.00 & -208.55$\pm$7.08 & \textbf{44.10$\pm$0.00} & -39.60$\pm$0.00 & -3.38$\pm$18.66 & 39.39$\pm$0.00 \\ \midrule[0.75pt]
GNN~\cite{GNN}  & -171.81$\pm$58.46 & -41.00$\pm$0.00 & -30.45$\pm$0.00 & -21.65$\pm$29.66 & -44.10$\pm$0.00 & -39.48$\pm$0.28 & 10.35$\pm$7.95 & -41.70$\pm$0.00 \\  \midrule[0.75pt]
Demo  & 9.91$\pm$0.62  & \textbf{41.00$\pm$0.00}        & -30.45$\pm$0.00  &  12.43$\pm$0.91 & \textbf{44.10$\pm$0.00}& -39.60$\pm$0.00 & 26.39$\pm$2.81 & 35.70$\pm$0.00 \\ 
DG-SAC (Ours)   & \textbf{25.90$\pm$10.78}   & \textbf{41.00$\pm$0.00}       & \textbf{18.57$\pm$0.27} & \textbf{38.96$\pm$18.10} & \textbf{44.10$\pm$0.00} & \textbf{44.13$\pm$0.10} & \textbf{33.01$\pm$0.00} & \textbf{41.46$\pm$0.48} \\ \bottomrule[1.25pt]
\end{tabular} }
\label{table:base}
\end{table*}

\subsection{Standard Baselines} \label{sec:exp:baseline}

We employ four state-of-the-art RL algorithms, namely Proximal Policy Optimisation (PPO)~\cite{ppo}, SAC, Deep Deterministic Policy Gradient (DDPG)~\cite{ddpg}, and Twin Delayed Deep Deterministic Policy Gradient (TD3)~\cite{td3} as the baselines for benchmarking our DG-SAC method. The results in Table~\ref{table:base} reveal that DG-SAC exhibits outstanding performance across the sub-tasks of various tasks, showcasing its effectiveness and adaptability in complex scenarios.

For the scooping sub-tasks (Spoon-Scoop and Scoop-Scoop), most baselines could scoop a small amount of granular material and tend to collide with the container. It was also found that these baselines tend to be trapped by local minima after collisions during training, incurring significant negative rewards and TCSs. 
On the other hand, although we design a reward function that encourages the agent to interact with particles, some baselines are found to be more affected by the penalties caused by the collisions with the container, and consequently lead to policies that keep lifting the tool and fail to complete the task. 

In the translating sub-tasks, none of the baselines have succeeded. We observe that these agents again tend to be trapped by local minima after colliding with the boundaries of the environment during training, leading to undesired behaviours. 

Lastly, in the pouring sub-tasks, the baselines generally perform well during training. They tend to quickly pour out all particles, thus resulting in high values of $\mathcal{R}_{\alpha}^{dist}$. However, these rapid pouring behaviours lead to low accuracy in directing the particles into the desired area. 

The poor performance of the RL baselines indicates the brittleness of training RL agents from scratch and the difficulty of reward design and pinpoints the effectiveness of demonstrations.

In addition to the RL baselines, we also select two common IL baselines that do not involve reward design: Behaviour Cloning (BC)~\cite{bc} and Generative Adversarial Imitation Learning (GAIL)~\cite{gail}. 
%As shown in Table~\ref{table:base}, the IL methods also perform worse than the DG-SAC method, likely due to their heavy reliance on the quality of the demonstrations (see Section~\ref{sec:exp:whyDGSAC}). 
As shown in Table~\ref{table:base}, the sixth and seventh rows (BC and GAIL) indicate that the performance of the IL methods is also inferior to that of the DG-SAC method, with the TCS mean values in all sub-tasks lower than those achieved by DG-SAC. This may be attributed to its heavy reliance on the quality of demonstrations (see Section~\ref{sec:exp:whyDGSAC}). Similar to the demonstrations, the policies learned by IL methods fail to pour granular materials in the pouring sub-tasks but perform relatively well in the translating tasks. Specifically, in the Scoop-Translate sub-task, both policies successfully translate all particles, achieving the same TCSs ($44.10$) as the DG-SAC model, and in the Spoon-Translate sub-task, they translate the majority of particles, resulting in a slightly lower TCS compared to DG-SAC. Notably in the scooping subtask, the trajectories generated by the BC method were comparable to, or even outperformed, those of the demonstrations. However, the GAIL method performed significantly worse in the scooping task, achieving very low training rewards and TCSs. Despite having a training time far exceeding that of BC, the policies generated by GAIL frequently collided with the container during execution.
%Similar to the demonstrations, the policies learned by IL methods fail to pour granular materials in the pouring sub-tasks but perform relatively well in the translating tasks. Notably, the trajectories generated by the BC method in the scooping sub-task do match or even surpass the performances of the demonstrations. However, the performance of GAIL in scooping tasks is significantly poor \zj{where xxxxx ... and xxxxx.}. Despite a training duration far exceeding that of BC, GAIL produces policies that frequently collide with the container. 

\subsection{Granular Manipulation-Specific Baseline} \label{sec:exp:gnn}

Beyond standard RL and IL-based baselines, we also evaluate a state-of-the-art GNN-based framework tailored to granular manipulation~\cite{GNN}, as the baseline. Prior work typically falls into two categories: real-world learning with physical feedback, which suffers from safety and efficiency issues, and simulation-based learning using surrogate models. As our method is simulator-driven by design, we focus on the latter for comparison. The GNN-based baseline~\cite{GNN} models particle dynamics using a learnt GNN and applies CMA-ES~\cite{cmaes} for control optimisation. It serves as a strong baseline, especially given the growing popularity of GNN-based methods in granular manipulation~\cite{Dynamic_GNN}.
%This method serves as a strong benchmark and is widely adopted in recent granular manipulation research.}

In our work, we generate training data using our simulator, which is configured to match the setup in~\cite{GNN}, also based on Taichi-MPM. The dataset covers all eight tasks. Linear perturbations are applied to the initial positions of the agent, granular material, and container, as well as to the action sequences, resulting in 1.8 million simulation frames. Then, rigid-body meshes are filled with particles to conform to the GNN input format. We follow the original network settings, where the model is trained for single-step prediction with 10 message-passing layers and 128-dimensional hidden states, with input features that include velocity, control sequences, and particle types. After a training of 430{,}000 steps, the learnt model is used for control optimisation with CMA-ES, with an initial variance of 1.5 for variables and population size of 20 for 150 iterations with five seeds. The results are shown in Table~\ref{table:base} under the GNN row.

As seen in Table~\ref{table:base}, the GNN performance is poor across all others, except for the bottle-related tasks, where the GNN baseline achieves marginal results. This highlights the fragility of the approach.
On the other hand, although both the GNN training and CMA-ES optimisation losses converge, the resulting trajectories are often physically invalid. Notably, in the GNN method~\cite{GNN}, it is trained and evaluated solely on a simple cup-pouring task, and, even then, the rendered trajectories exhibit frequent violations, such as particles penetrating solid boundaries. When deployed to different tasks in our work, these shortcomings are further amplified. This suggests the model fails to capture underlying dynamics and, instead, overfits the training distribution. Consequently, CMA-ES often converges to behaviours that are numerically stable but physically implausible or task-irrelevant. A further limitation lies in the data-driven nature of the method. Granular manipulation involves high-dimensional, continuous action spaces, and even our large datasets fail to cover the space of relevant trajectories, resulting in poor generalisation. Moreover, collecting such datasets is costly in both manual effort and computational resources.

In contrast, our method avoids these issues by directly leveraging the simulator, eliminating the need for offline model training or manual data collection. More importantly, it enables accurate, physically consistent trajectory optimisation without relying on learnt approximations. Consequently, the GNN-based baseline fails to produce competitive results after training for more than four days, while our approach is significantly more efficient, accurate, and consistently successful across all tasks.

\subsection{Ablation Studies} \label{sec:exp:ablation}
This section presents the ablation studies on the DG-SAC agent first to examine the impact of physical information on the training process. Four scenarios related to physical information were considered: 
\begin{itemize}
    \item using only particle position information as observations;
    \item using only particle velocity information as observations;
    \item excluding particle observational physical information from observations;
    \item not downsampling the observed physical information of the granular materials.
\end{itemize}
Secondly, we perform ablation experiments on the structure of our model to validate the impact of incorporating the demonstration replay buffer during training. For clarity, we define this model as DGN-SAC, indicating a policy model in which demonstrations influence network parameter weights only during the demonstration adding process, without sampling from the additional replay buffer during training process.

Experimental results in Table~\ref{table:ablation} indicate that the full agent with particle position and velocity information as observations consistently achieves the best performances in most tasks (6/7), with only a small gap to the optimal result in the remaining task. Position information is shown to play a critical role during training, with trajectories trained using position data generally outperforming others. This is particularly evident in the relatively less challenging pouring sub-tasks, where the "Pos Only" models achieve performance comparable to the full agent and even slightly outperform it in the sub-task using the bottle. Similarly, it exhibits strong performance in the scoop-translating sub-task but is slightly less effective in the more challenging spoon-translating sub-task (which is intuitive, as using a spoon is more prone to volume loss). However, in the scooping task, position information alone seems to be insufficient. Although it can complete the task in some trials of the scoop-scooping sub-task, it fails to avoid collisions with the container in others, resulting in low TCSs. In other cases, it tends to converge to local optima that avoid collisions but fail to complete the task. These results show that using only position information can yield satisfactory results in simpler tasks but tends to be weaker and less stable as task complexity increases. 

\begin{table*}[tp!]
%\vspace{0.3cm}
\centering
\caption{The training reward (top) and task completion score (bottom) with their standard deviations for ablation studies.}
\resizebox{1.8\columnwidth}{!}{
\begin{tabular}{c|ccc|ccc|c|c}
\toprule[1.25pt]
Task     & Spoon-Scoop & Spoon-Translate & Spoon-Pour & Scoop-Scoop & Scoop-Translate & Scoop-Pour & Bottle-Pick\&Pour & Shovel-Move \\ \midrule[0.75pt]
Pos Only  & -51.81$\pm$1.57  &  35.80$\pm$0.11  &  94.32$\pm$24.42 & 82.71$\pm$72.47 &   53.61$\pm$24.08 &  \textbf{98.25$\pm$4.41} & \textbf{106.82$\pm$7.93} & 70.98$\pm$2.36 \\
Vel Only   & -130.67$\pm$5.85  &  -64.34$\pm$2.61  &  94.11$\pm$0.53 & -69.41$\pm$10.74 &   -70.40$\pm$6.64 & 89.62$\pm$2.76 & 102.96$\pm$20.78 & -57.86$\pm$27.38 \\ 
No Obs Info    & -138.82$\pm$0.08  &  14.73$\pm$19.58  &  \textbf{95.73$\pm$0.09} & -27.09$\pm$18.61 &   39.03$\pm$15.02 & -61.95$\pm$3.95 & 1.66$\pm$33.63 & 30.69$\pm$54.49 \\ 
No Downsampling   & -51.42$\pm$0.84  &  ---  & --- & -24.57$\pm$2.65 & --- & ---& 98.89$\pm$0.01 & --- \\ \midrule[0.75pt]
DGN-SAC  & 39.79$\pm$23.45   & 3.12$\pm$24.29      & -40.77$\pm$14.88  &  24.84$\pm$16.79 &   31.78$\pm$4.61 & 51.16$\pm$19.47 & 104.71$\pm$4.72 & -29.59$\pm$62.23 \\ \midrule[0.75pt]
DG-SAC (Ours)   & \textbf{49.87$\pm$20.95}   & \textbf{60.01$\pm$0.05}       & 49.93$\pm$6.32 & \textbf{93.13$\pm$35.39} & \textbf{86.96$\pm$4.69}& 63.91$\pm$7.20 & 98.79$\pm$0.00 & \textbf{74.75$\pm$2.66} \\ \bottomrule[1.25pt]
\toprule[1.25pt]
Task     & Spoon-Scoop & Spoon-Translate & Spoon-Pour & Scoop-Scoop & Scoop-Translate & Scoop-Pour & Bottle-Pick\&Pour & Shovel-Move \\  \midrule[0.75pt]
Pos Only  &  -12.51$\pm$0.14 &  13.88$\pm$0.48  &   17.85$\pm$0.79 & 17.41$\pm$31.24 &   35.28$\pm$12.47 & 42.19$\pm$2.37 & \textbf{33.15$\pm$0.03} & 41.22$\pm$0.25 \\
Vel Only   &  -127.08$\pm$5.84 &  -39.22$\pm$1.78  &  8.12$\pm$0.02 & -57.89$\pm$5.50 &   -44.10$\pm$0.00 &  41.72$\pm$1.77 & 13.91$\pm$10.00 & -30.66$\pm$11.16 \\ 
No Obs Info    &  -129.79$\pm$0.07 &  3.73$\pm$15.83  &  8.11$\pm$0.02 & -23.27$\pm$15.96 &   15.42$\pm$11.88 &  -39.01$\pm$0.59 & 10.36$\pm$2.80 & 19.64$\pm$26.58 \\ 
No Downsampling    & -13.03$\pm$0.05  &  ---  & --- &  -12.25$\pm$0.90 & --- &  --- & 33.01$\pm$0.01 & --- \\ \midrule[0.75pt]
DGN-SAC  & 20.51$\pm$6.12   & -8.90$\pm$19.00       & -13.31$\pm$8.66 &  -1.20$\pm$4.00 &   15.38$\pm$4.39 & 31.88$\pm$10.26 & 33.00$\pm$0.17 & -23.48$\pm$32.73 \\ \midrule[0.75pt]
DG-SAC (Ours)   & \textbf{25.90$\pm$10.78}   & \textbf{41.00$\pm$0.00}       & \textbf{18.57$\pm$0.27} & \textbf{38.96$\pm$18.10} & \textbf{44.10$\pm$0.00} & \textbf{44.13$\pm$0.10} & 33.01$\pm$0.00 & \textbf{41.46$\pm$0.48} \\ \bottomrule[1.25pt]
\end{tabular} }
\label{table:ablation}
\end{table*}

Secondly, the results show that using only velocity as observations or omitting physical information entirely leads to significantly poor performances. This issue is particularly pronounced in the scooping tasks, where frequent collisions with the container result in low training rewards and TCSs. 

Thirdly, the down-sampling operation proves highly effective when the particle count is large, significantly improving the training speed ($4.67 \times$ with a spoon, $5.33 \times$ with a scoop, and $1.27 \times$ with a bottle) while yielding superior training rewards and TCSs. In scenarios with fewer particles, such as the bottle-related sub-task, down-sampling has minimal impact on the performance, primarily serving to enhance training speed. In summary, the results show that using both position and velocity information as observations is crucial for effective physics-informed learning, and the down-sampling operation is crucial for efficient training.

Finally, the results also reveal that DGN-SAC generally performs poorer compared to the full agent. This suggests that allowing demonstration trajectories to continuously influence network weights through sampling from the additional replay buffer can effectively enhance training outcomes.

\subsection{Skill Chaining} \label{sec:exp:chaining}

After analysing the performance of each sub-task, this subsection examines the transitions between sub-tasks, as these are crucial for the successful completion of the overall task. To validate the effectiveness of our designed skill chaining structure, we present in Table~\ref{table:euler} the variations in the Euler angles of the agent of our model at the last timestep of the scooping sub-tasks over five trials under the stimulus of reward $\mathcal{J}_{\alpha}$. The agent is expected to achieve a change of $-110^{\circ}$ in $\vartheta^{z}$ before the completion of scooping to facilitate a better transition to the translating sub-task. 

Experimental results indicate that our learning framework can seamlessly integrate sub-tasks requiring specialised transitions with relatively small errors. Notably, the relative error in skill chain transitions for the spoon-based task is lower than that for the scoop-based task. The reason for this outcome lies in the higher weighting in $\mathcal{J}_{\alpha}$ within the reward function $\mathcal{R}_{s}$ for the spoon-scooping sub-task. Overall, our skill chaining structure effectively integrates the scooping and transporting sub-tasks across different tools, thereby validating the efficacy of our skill chain approach.
\begin{table}[h]
%\vspace{0.3cm}
\centering
\caption{The variations of the Euler angle of the agent and their relative errors to the target value at the transition between the scooping and translating sub-tasks.}
\resizebox{1.0\columnwidth}{!}{
\begin{tabular}{c|ccc}
\toprule[1.25pt]
Task  & Target & Result & Relative Error \\ \midrule[0.75pt]
Spoon-Scoop & -110$^{\circ}$   & -113.03$^{\circ}$$\pm$2.23$^{\circ}$    & 2.75\%$\pm$2.03\% \\ 
Scoop-Scoop & -110$^{\circ}$   & -119.10$^{\circ}$$\pm$5.50$^{\circ}$    & 8.27\%$\pm$5.00\%            \\\bottomrule[1.25pt]
\end{tabular} }
\label{table:euler}
\end{table}

\subsection{Beyond Fluid-like Demonstrations}
\label{sec:exp:elastic}

Beyond the seven granular manipulation tasks, we further evaluate the generalisability of our framework in more challenging kitchen scenarios, where fluid-like materials are hard to control, and gradient-based optimisation often fails to produce viable trajectories. Tasks like Shovel-Move—transporting granular material across a flat surface are particularly difficult for fluid models, as fluids tend to spread uncontrollably, making them unsuitable for directed transport. Consequently, using fluid simulation to generate effective demonstrations in such contexts is impractical.

To address this, we replace the fluid model with an elasto-plastic material model, which offers greater stability and controllability than the DP model used for granular media. We simulate the elasto-plastic behaviour using the von Mises yield criterion~\cite{jones2009deformation} with a yield stress of $10$, producing clay-like dynamics amenable to gradient-based optimisation. As before, the resulting trajectories are used as demonstrations. The rightmost columns of Tables~\ref{table:base} and~\ref{table:ablation} show the results of the Shovel-Move task. Our method consistently achieves the best performance, in contrast to standard RL methods, which occasionally reach high TCS but remain unstable across trials. These results highlight the generalisability, robustness and effectiveness of our method in difficult manipulation settings.
\subsection{Generalisation to Varying Material Properties}
\label{sec:exp:material}
\begin{table*}[htp!]
%\vspace{0.3cm}
\centering
\caption{Robustness of trajectory performance across granular materials with varying physical parameters.}
\resizebox{1.8\columnwidth}{!}{
\begin{tabular}{c|ccc|ccc|ccc|ccc|c}
\toprule[1.25pt]
Parameter Combination & \multicolumn{3}{c|}{Spoon-Scoop} & \multicolumn{3}{c|}{Spoon-Translate} & \multicolumn{3}{c|}{Spoon-Pour} & \multicolumn{3}{c|}{Bottle-Pick\&Pour} & \multirow{2}{*}{Remark} \\ 
$E/\nu/\phi_{f}$ & TD & Reward & TCS & TD & Reward & TCS & TD & Reward & TCS & TD & Reward & TCS & \\  \midrule[0.75pt]
$1000/0.2/30^{\circ}$  & ---  &  57.23  &  35.27  &  --- &  60.00  & 41.00 & ---  & 50.65 &  18.87 & ---  & 96.61 & 32.93 & Baseline \\
$800/0.2/30^{\circ}$  & 7.37e-8  &  57.38  &  35.27  & 3.48e-8 & 60.03 & 41.00 & 5.78e-3 & 59.52 & 17.92 & 5.58e-5 & 96.69 & 33.07 & Softer \\
$900/0.2/30^{\circ}$  & 2.00e-8  &  57.31 &  35.27 &  7.77e-9 & 60.01 & 41.00 & 2.39e-3  & 55.11 & 18.89 & 2.66e-5 & 96.66 & 33.00 & Softer \\
$1100/0.2/30^{\circ}$  & 1.63e-7 & 57.03 &  35.37  &  5.21e-9 & 59.99 & 41.00 & 1.88e-3 & 47.77 & 19.32 & 2.58e-5 & 96.27 & 32.55 & Stiffer \\
$1200/0.2/30^{\circ}$  & 1.24e-7  & 57.21  & 35.47  &  1.88e-8 &  59.98  & 41.00 & 3.25e-3  & 43.52 & 19.04 & 4.84e-5 & 96.81 & 33.08 & Stiffer \\
$1000/0.1/30^{\circ}$  & 2.10e-7  &  57.31  &  35.27  &  3.40e-9 &  59.99  &  41.00 & 4.00e-4  & 49.20 & 18.66 & 8.80e-5 & 96.11 & 32.68 & More compressible \\
$1000/0.15/30^{\circ}$  & 3.70e-8  &  57.37  &  35.17  &  7.88e-10 & 59.99  &  41.00 & 3.16e-4  & 50.48 & 19.18 & 4.12e-5 & 96.82 & 33.25 & More compressible\\
$1000/0.25/30^{\circ}$  & 5.16e-8  &  56.73  &  35.17  &  7.99e-10 &  60.00  &  41.00 & 3.45e-4 & 50.67 & 18.76 & 3.99e-5 & 97.14 & 33.35 & Less compressible \\
$1000/0.3/30^{\circ}$  & 1.52e-7  &  56.06  &  34.87  & 2.89e-9 &  60.01  &  41.00 & 1.21e-3 & 48.49 & 18.69 & 7.75e-5 & 96.92 & 33.05 & Less compressible \\
$1000/0.2/20^{\circ}$  & 7.40e-12  &  57.22  &  35.27  & 2.10e-7 &  60.08  &  41.00 & 2.68e-3 & 45.05 & 19.01 & 7.71e-5 & 96.17 & 32.78 & Slipperier \\
$1000/0.2/25^{\circ}$  & 6.49e-12  &  57.20  &  35.27  & 3.84e-8 &  60.04  &  41.00 & 1.75e-3 & 46.57 & 18.68 &  3.27e-5 & 96.02  & 32.50 & Slipperier \\
$1000/0.2/35^{\circ}$  & 7.36e-12  &  57.21  &  35.27  & 3.42e-8 &  59.96  &  41.00 & 1.18e-3  & 53.53 & 18.83 & 3.31e-5 & 97.11 & 33.25 & Stickier \\
$1000/0.2/40^{\circ}$  & 4.67e-12  &  57.22  &  35.27  & 1.23e-7 &  59.93  &  41.00 & 3.09e-3  &  57.47 & 18.69 & 7.81e-5 & 96.61 & 32.57 & Stickier \\
 \bottomrule[1.25pt]
\toprule[1.25pt]
Parameter Combination & \multicolumn{3}{c|}{Scoop-Scoop} & \multicolumn{3}{c|}{Scoop-Translate} & \multicolumn{3}{c|}{Scoop-Pour} & \multicolumn{3}{c|}{Shovel-Move} & \multirow{2}{*}{Remark} \\ 
$E/\nu/\phi_{f}$ & TD & Reward & TCS & TD & Reward & TCS & TD & Reward & TCS & TD & Reward & TCS & \\  \midrule[0.75pt]
$1000/0.2/30^{\circ}$  & ---  & 115.45 & 45.05  &  ---& 86.75 &  44.10 & ---  &  60.49 & 42.75 & ---  & 69.72 & 41.70 & Baseline \\
$800/0.2/30^{\circ}$  & 1.04e-7 & 117.72 & 45.95 & 8.10e-7 & 86.70  & 44.10 & 3.29e-3 & 62.18 & 43.00 & 4.55e-4 & 52.60 & 27.90 & Softer \\
$900/0.2/30^{\circ}$  & 2.59e-8 & 116.53 & 45.45 & 2.46e-8 & 86.72 & 44.10 & 1.38e-3 & 62.59 & 44.25 &  2.05e-4 & 57.96  & 32.10 & Softer \\
$1100/0.2/30^{\circ}$  & 1.64e-8 & 115.23 & 45.05 & 2.01e-8 & 86.78 &  44.10 & 1.55e-3 & 63.30 & 45.00 & 2.09e-4 & 68.58  & 41.70 & Stiffer \\
$1200/0.2/30^{\circ}$  & 5.56e-8  & 115.71 & 45.35 & 7.62e-8 & 86.80 &  44.10 & 1.25e-3 & 62.67 & 44.50 & 5.07e-4 & 67.06 & 41.70 & Stiffer \\
$1000/0.1/30^{\circ}$  & 1.75e-7 & 115.71 & 45.35 & 5.88e-9 &  86.72  &  44.10 & 1.38e-3 &  61.52 & 43.25 & 3.00e-5 & 69.23 & 41.70 & More compressible \\
$1000/0.15/30^{\circ}$  & 3.53e-8 & 115.46 & 45.05 & 1.64e-9 & 86.73  &  44.10 & 1.35e-3 & 61.34 & 43.00 &  4.87e-7 & 69.74 & 41.70 & More compressible\\
$1000/0.25/30^{\circ}$  & 2.80e-8 & 115.73 & 45.35 & 2.00e-9 & 86.77  &  44.10 & 5.55e-4 & 59.52 & 41.50 &  1.29e-5 & 69.05 & 41.70 & Less compressible \\
$1000/0.3/30^{\circ}$  & 1.02e-7 & 115.91 & 45.85 & 9.36e-9 & 86.78  &  44.10 & 8.07e-4  & 60.59 & 42.50 & 1.34e-5 & 69.02 & 41.70 & Less compressible \\
$1000/0.2/20^{\circ}$  & 4.63e-5  & 103.28 & 42.35 & 1.32e-7 & 86.81 &  44.10 & 1.99e-3 &  62.91 & 44.25 &  3.47e-4 & 59.17 & 33.30 & Slipperier\\
$1000/0.2/25^{\circ}$  & 9.19e-7 & 109.41 & 43.75 & 3.05e-8 & 86.78  &  44.10 & 6.29e-4 &  60.27 & 42.25& 2.76e-4 & 61.61 & 35.70 & Slipperier \\
$1000/0.2/35^{\circ}$  & 1.01e-6  &  124.26  &  48.45  & 2.52e-8 & 86.74  & 44.10 & 1.35e-3 & 61.26 & 43.00 & 4.61e-4 & 67.34 & 41.70 & Stickier \\
$1000/0.2/40^{\circ}$  & 2.18e-5  & 128.53 &  49.95  & 9.18e-8 &  86.73  & 44.10 & 2.08e-3 &  61.17 & 42.75 &  5.87e-4 & 67.54 & 41.70 & Stickier \\
\bottomrule[1.25pt]
\end{tabular} 
}
\label{table:metarials}
\end{table*}
To assess the robustness of our approach under varying physical conditions, we evaluate the optimised trajectories on granular materials with diverse mechanical properties. Real-world substances like flour, sugar, and salt differ in stiffness, compressibility, and friction, which we model by varying the Young’s modulus $E$, Poisson’s ratio $\nu$, and friction angle $\phi_f$. Lower values of these parameters correspond to softer, more compressible, and more slippery materials, while higher values indicate stiffer, less compressible, and more adhesive behaviours. All tested parameters remain within meaningful ranges, for example, overly low $E$ yields fluid-like behaviours, while overly high $E$ leads to rigid-body-like dynamics.

We randomly sample optimised trajectories from our training runs and test them across environments with different combinations of $E$, $\nu$, and $\phi_f$, covering a broad range of realistic material properties. For each configuration, we report task reward, TCS, and trajectory deviation (TD), defined as the mean per particle positional deviation under altered material conditions to quantify the sensitivity to physical variation. Table~\ref{table:metarials} summarises the performance across multiple settings. Since granular manipulation depends on stable tool-material interactions, lower stiffness or friction often results in more dispersed behaviours. Nonetheless, the trajectories remain effective, demonstrating the robustness of our method across material variations.
\subsection{Unsuccessful Trials with Set-Based Critic Representations} \label{sec:exp:pointnet} 

To further evaluate the generalisation of our approach, we tested a PointNet-SAC~\cite{pointnet} variant, where the granular material is encoded as an unordered point cloud for the critic. However, meaningful policies emerged on only two sub-tasks, and results are inconsistent across five trials.

We attribute this failure to the PointNet’s architecture, which processes each particle independently and aggregates features via global max pooling. This design neglects local structural dependencies crucial for modelling inter-particle interactions, resulting in a degraded observation embedding that lacks relational information, ultimately impairing the critic's ability to support policy learning.

Although the experiment did not yield usable outcomes, it underscores a key insight: our method is not bound to a specific RL backbone, e.g., SAC, and can generalise across learning frameworks, provided the particle representation preserves meaningful local structures.

\subsection{Real-World Manipulation} \label{sec:exp:real}

\begin{figure}[htb!]
   % \vspace{-0.1cm} 
    \centering
    \includegraphics[width=8cm]{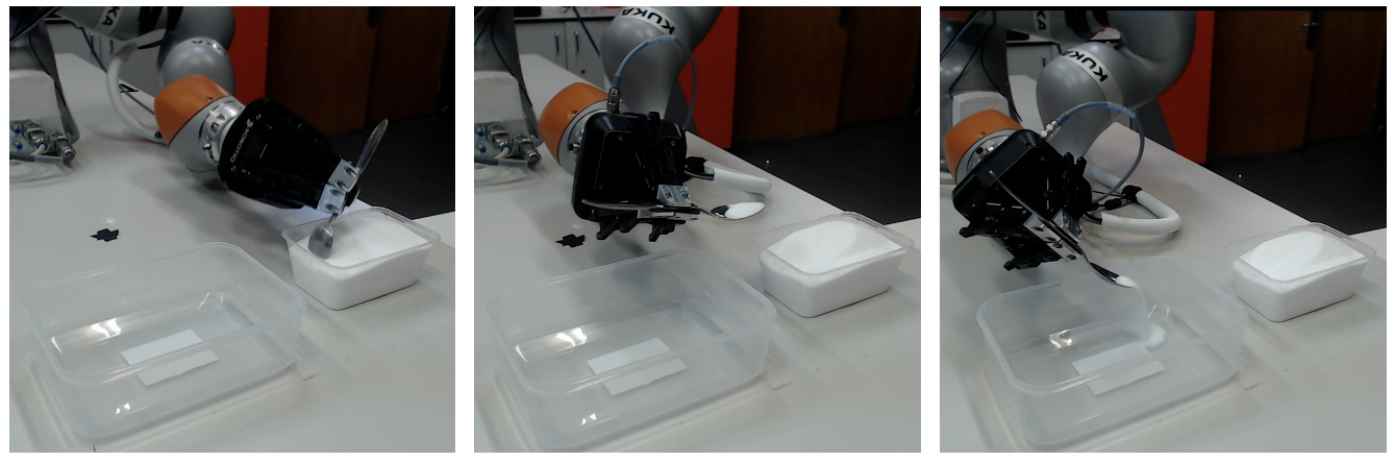}
    \caption{A sequence of snapshots showing that our real robot successfully completed the spoon-based task of transporting granular materials between two containers.}
    \label{fig:5}
    %\vspace{-0.2cm} 
\end{figure}

To verify the performance of our method in real-world environments, we transfer the policy to a real robot. As shown in Fig.~\ref{fig:5}, we use a seven-degree-of-freedom robot manipulator, Kuka lbr iiwa (14kg) equipped with a ROBOTIQ 3-finger robot gripper. To demonstrate that our method remains effective across real-world scenarios involving varying physical conditions, we select granular materials, including salt, sugar, and flour, with distinct properties, in terms of cohesion and flow behaviours, highlighting its robustness beyond a single setup. The experimental setup comprises a container for storing granular materials and another one to serve as the target zone. Given the inherent properties of granular materials, which complicate quantitative analysis in real-world settings, we define the criterion for task completion as the visually confirmed transfer of granular materials into the target container without colliding with either container. We conducted tests using the optimal trajectories obtained from our simulated environment and undertook three experiments for the three types of granular materials mentioned above. 

It was found that the robot accomplished all tasks in all trials without colliding with the container. This highlights the effectiveness and feasibility of transferring the skills learnt by our method from simulated to real-world settings, and also reflects the close alignment between our simulation environment and the real-world scenario.

\section{CONCLUSIONS} \label{sec:con}

This paper proposes a novel physics-informed learning framework for granular material manipulation. A differentiable simulator is built based on the MLS-MPM and the DP yield model for simulating four complex, long-term granular manipulation tasks, comprising eight sub-tasks. We propose to fill the static rigid mesh (container) with static particles in the simulation to facilitate more efficient and accurate collision detection. To accelerate training, DG-SAC leverages demonstrations optimised in materials with smooth or continuous dynamics, such as fluids or elasto-plastic solids, while targeting sub-tasks involving granular materials, whose gradients are unstable and ill-suited for direct optimisation. In addition, it benefits from careful reward design, including an important skill chaining reward that enables smooth transitions between consecutive sub-tasks.

Experimental results show that DG-SAC, combined with position and velocity-based physical information, outperforms several baselines and accomplishes the proposed complex tasks in simulation and the real world with low variance across multiple runs. These results further validate the effectiveness of our method in consistently achieving superior performance while demonstrating flexibility and robustness across diverse tasks and varying material properties.

In the future, we intend to expand our learning framework to a broader range of application scenarios, such as granular material manipulation in contexts like gardening and beach environments, rather than being limited to the four tasks in kitchen-related scenarios explored in this study. Another research direction focuses on addressing the gradient explosion problem in granular material manipulation from its principles, aiming to enable more stable and efficient gradient-based optimisation in contact-rich tasks.

\bibliographystyle{ieeetr}
\bibliography{IEEEexample}

\begin{IEEEbiography}[{\includegraphics[width=1in,height=1.25in,clip,keepaspectratio]{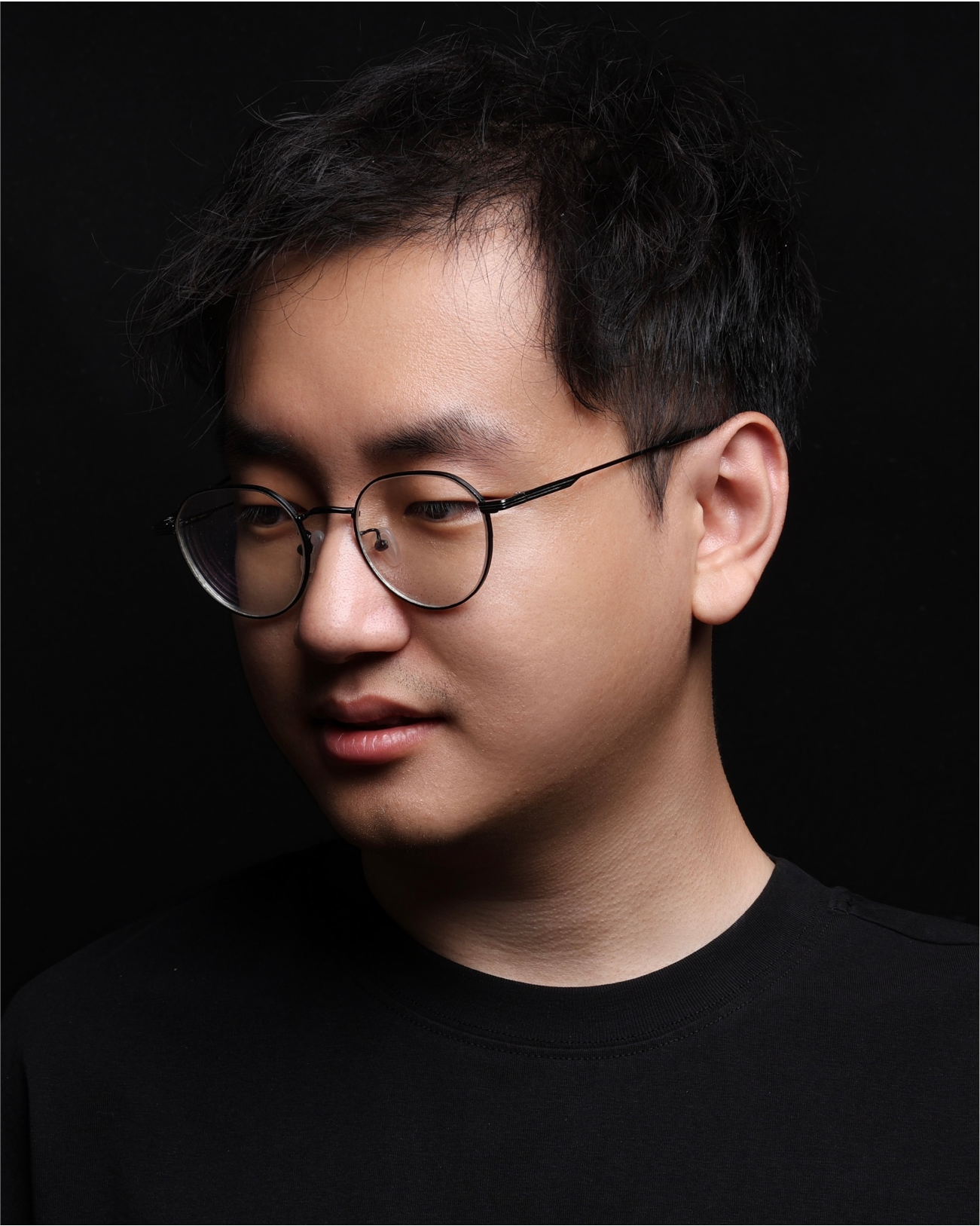}}]{Minglun Wei} (Student Member, IEEE) received the B.Eng. degree in Microelectronics from Northwestern Polytechnical University, Xi’an, China, in 2020, and the M.Sc. degree in Signal Processing and Communications from The University of Edinburgh, U.K., in 2021. He is currently pursuing the Ph.D. degree with Cardiff University, Cardiff, U.K.

Prior to starting his Ph.D. studies, he worked in industry on projects applying large language models to AI for Science. His research interests include robotic manipulation of deformable objects and data-driven modelling of dynamical systems.
\end{IEEEbiography}

\begin{IEEEbiography}[{\includegraphics[width=1in,height=1.25in,clip,keepaspectratio]{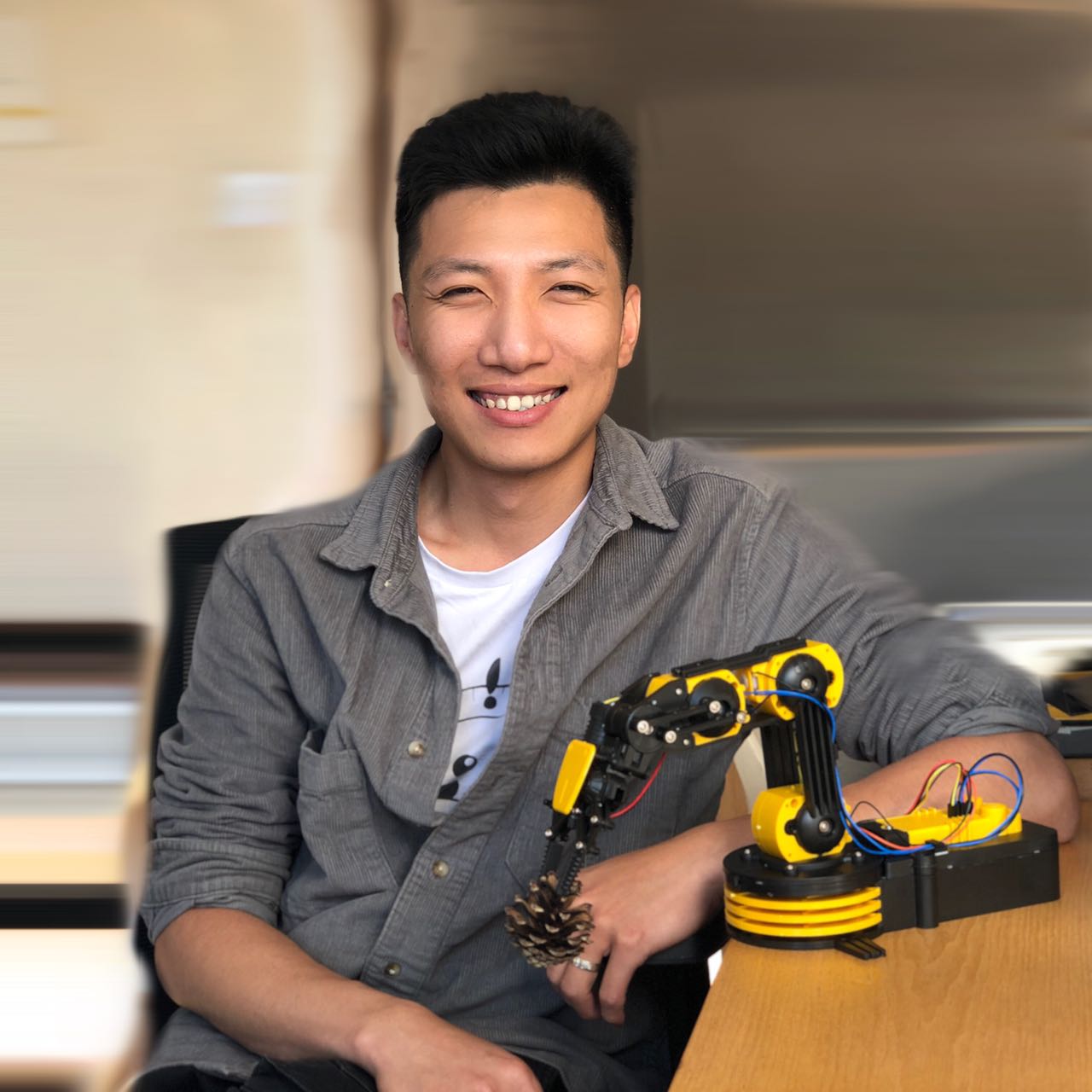}}]{Xintong Yang} received his Ph.D. from Cardiff University, Cardiff, U.K., in 2023, and his Bachelor’s and Master’s degrees in Mechanical and Industrial Engineering from Guangdong University of Technology, Guangzhou, China, in 2016 and 2019. He has been a research associate (postdoc) in the School of Engineering at Cardiff University since Jan. 2023. He specialised in the robotic manipulation of real-world objects, rigid or deformable, through model-based and/or data-driven methods. Currently, he is primarily responsible for developing a robotic platform for automatically conducting biology/chemical experiments for AI-driven science discovery. He is also working on developing a real-world-applicable robotic manipulation system.
\end{IEEEbiography}

\begin{IEEEbiography}[{\includegraphics[width=1in,height=1.25in,clip,keepaspectratio]{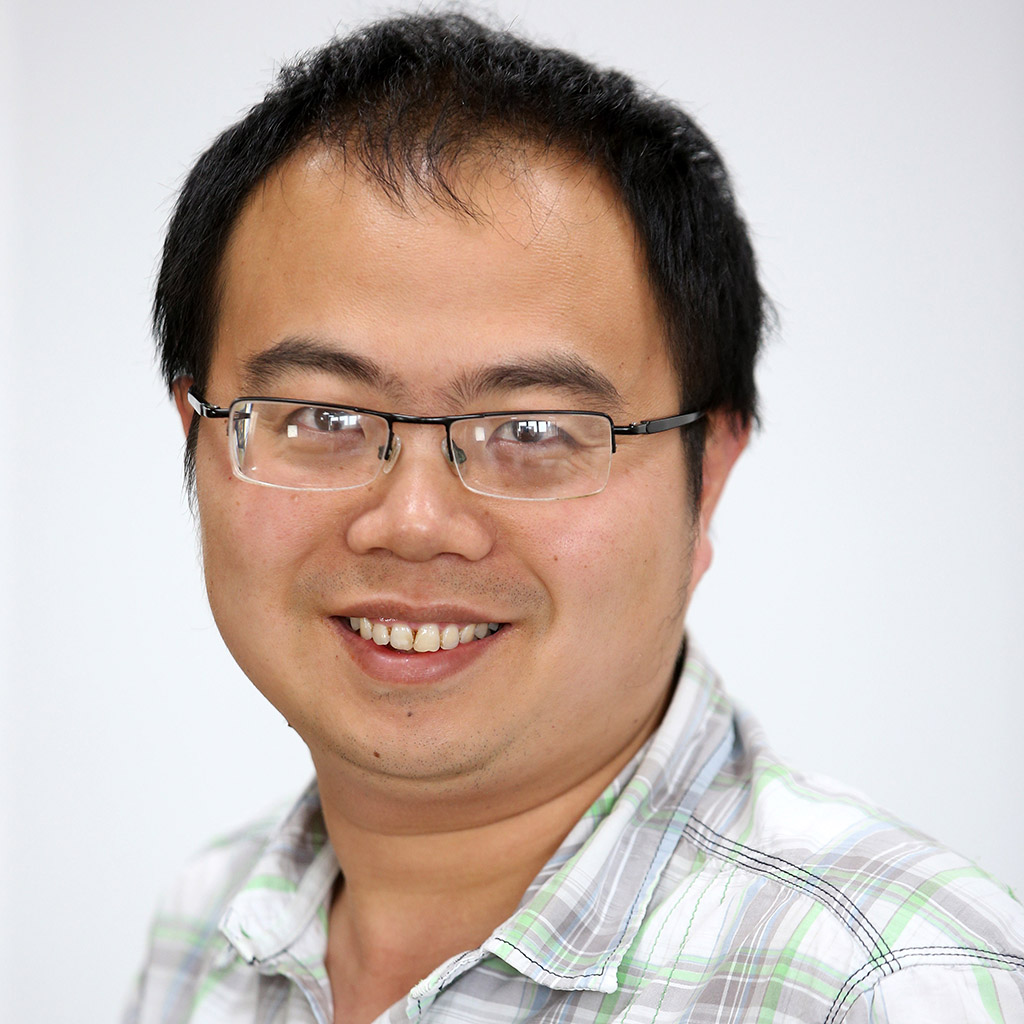}}]{Yu-Kun Lai} (Senior Member, IEEE) received his bachelor’s and PhD degrees in computer science from Tsinghua University, in 2003 and 2008, respectively. He is currently a professor in the School of Computer Science \& Informatics, Cardiff University, UK. His research interests include computer graphics, geometry processing, image processing, and computer vision. He is on the editorial boards of IEEE Transactions on Visualization and Computer Graphics, Computers \& Graphics, and The Visual Computer.
\end{IEEEbiography}

\begin{IEEEbiography}[{\includegraphics[width=1in,height=1.25in,clip,keepaspectratio]{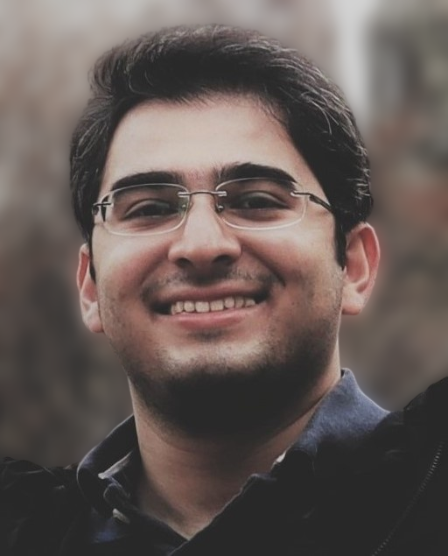}}]{Seyed Amir Tafrishi} (Member, IEEE) received his M.Sc. degree in control systems engineering from the University of Sheffield in 2014, UK, and Ph.D. degree in mechanical engineering from Kyushu University, Japan in 2021. 

Dr. Tafrishi is currently a Lecturer at Engineering School, Cardiff University, UK. He is the founder and head of the Geometric Mechanics and Mechatronics in Robotics (gm$^2$R) lab. He was a Specially Appointed Assistant Professor working on Mooonshot R \& D project JST at Tohoku University, Japan, between 2021-2022. Since 2014, he has been a visiting researcher at University of Sheffield, the Mechatronics Lab at METU, Turkey, and the Fluid Mechanics Lab at the University of Tabriz, Iran. His research interests include robotics, mechanism design, reconfigurable robots, rolling contact, geometric mechanics, under-actuated systems.
\end{IEEEbiography}

\begin{IEEEbiography}[{\includegraphics[width=1in,height=1.25in,clip,keepaspectratio]{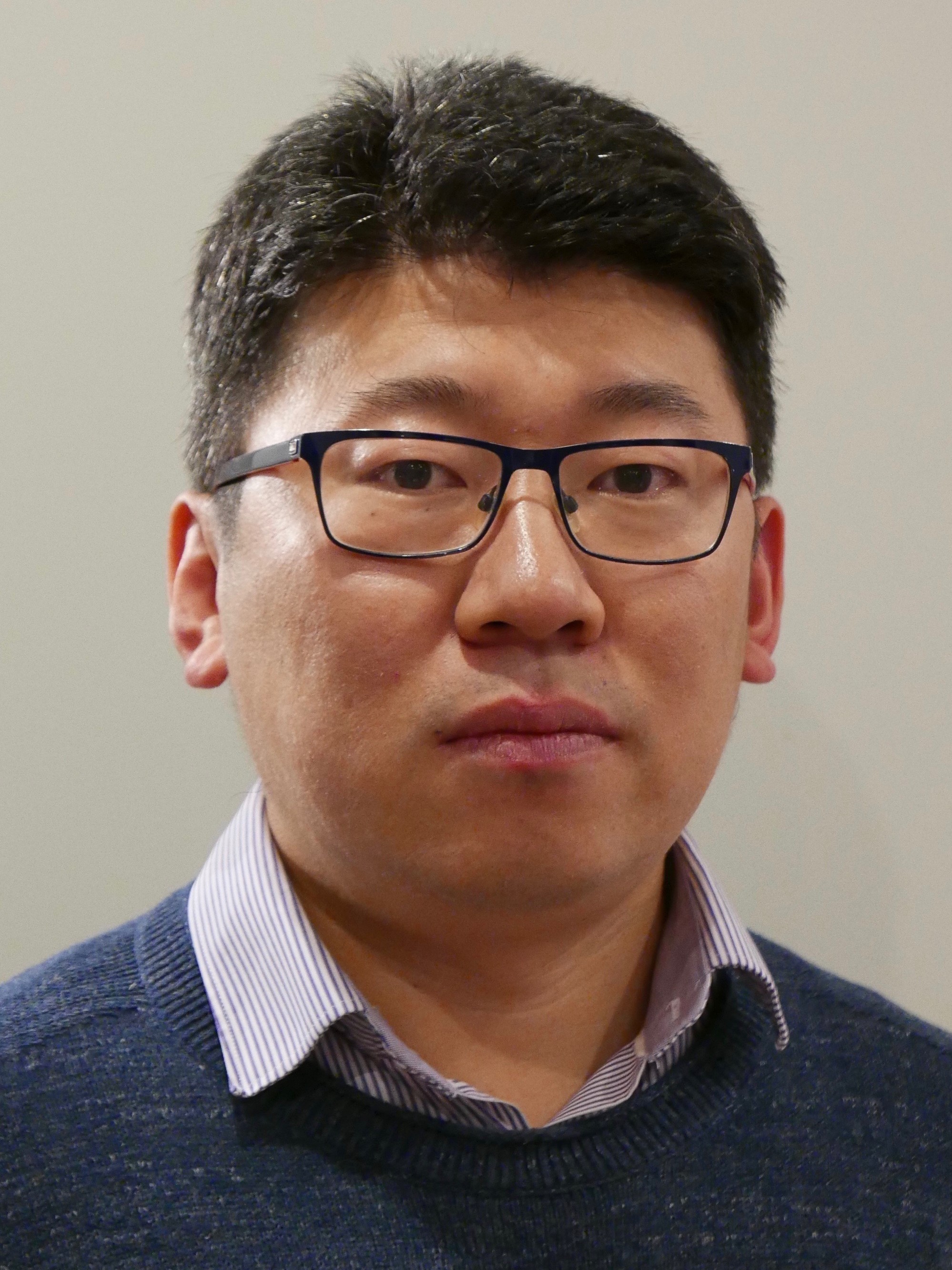}}]{Ze Ji} (Member, IEEE) received the Ph.D. degree from Cardiff University, Cardiff, U.K., in 2007. He is a Reader with the School of Engineering, Cardiff University, UK. Prior to his current position, he was working in industry (Dyson, Lenovo, etc) on autonomous robotics. His research interests are broad, including robot manipulation, robot learning, autonomous robot navigation, physics-informed learning, computer vision, simultaneous localization and mapping (SLAM), acoustic localization, and tactile sensing. He is on the editorial boards of several journals, including IEEE/ASME Transactions on Mechatronics.
\end{IEEEbiography}
 
\end{document}